\documentclass[10pt,onecolumn,letterpaper]{article}
\usepackage{cvpr}
\usepackage[numbers,sort&compress]{natbib}
\usepackage{times}
\usepackage{eso-pic}

\usepackage{epsfig}
\usepackage{graphicx}
\usepackage{amsthm}
\usepackage{amsmath}
\usepackage{amssymb}
\usepackage[breaklinks=true,bookmarks=false]{hyperref}

\numberwithin{equation}{section}
\usepackage{datetime}

\usepackage[utf8]{inputenc} % allow utf-8 input
\usepackage[T1]{fontenc}    % use 8-bit T1 fonts
\usepackage{hyperref}       % hyperlinks
\usepackage{url}            % simple URL typesetting
\usepackage{booktabs}       % professional-quality tables
\usepackage{amsfonts}       % blackboard math symbols
\usepackage{nicefrac}       % compact symbols for 1/2, etc.
\usepackage{microtype}      % microtypography
\usepackage{xcolor}         % colors

\usepackage{graphicx}
\usepackage{amsmath} 
\usepackage{algorithm}
\usepackage{algorithmic}
\usepackage{colortbl} 
\usepackage{tcolorbox}
\tcbuselibrary{skins,breakable}

\usepackage{multirow}
\usepackage{array}
\usepackage{longtable}
\usepackage{adjustbox}
\usepackage{array}
\usepackage{lscape}
\usepackage{natbib}
\usepackage[a4paper,margin=1in,headheight=15pt,headsep=24pt]{geometry}

\usepackage{fancyhdr}

\usepackage{amsmath,amsfonts,bm}

\def\eqref#1{equation~\ref{#1}}
\def\1{\bm{1}}

\DeclareMathAlphabet{\mathsfit}{\encodingdefault}{\sfdefault}{m}{sl}
\SetMathAlphabet{\mathsfit}{bold}{\encodingdefault}{\sfdefault}{bx}{n}

\usepackage{hyperref}
\usepackage{url}
\usepackage{graphicx}
\usepackage{multirow}
\usepackage{booktabs}
\usepackage{array}
\usepackage[table]{xcolor}
\usepackage{pifont}
\usepackage{subcaption}

\usepackage{algorithm}
\usepackage{algorithmic}
\usepackage{amsmath,bm}
\usepackage{amssymb}
\usepackage{amsthm}
\usepackage{multirow} 
\usepackage{graphicx}
\usepackage{amsmath}
\usepackage{amsthm}

\usepackage{enumitem}
\usepackage{float}
\usepackage{placeins}

\cvprfinalcopy % *** Uncomment this line for the final submission
\def\cvprPaperID{****} % *** Enter the CVPR Paper ID here
\allowdisplaybreaks
\newcommand{\rsubhead}[1]{\par\addvspace{0.6em}\noindent\textbf{#1}}

\begin{document}
\lhead{}
\lfoot{\date{\today},\date{\currenttime}}
\rfoot{NGD for DL}

% \title{Instruction Learning Paradigms: A Dual Perspective on White-box and Black-box LLMs}
\title{FBOS-RL: Feedback-Driven Bi-Objective Synergistic Reinforcement Learning}
\author{
% Yanwei Ren, Liu Liu, Baosheng Yu, Jiayan Qiu, Quan Chen
\textbf{Xikai Zhang}$^{1}$\quad
\textbf{Yongzhi Li}$^{3}$\quad
\textbf{Likang Xiao}$^{1}$\quad
\textbf{Yingze Zhang}$^{1}$\quad
\textbf{Yanhua Cheng}$^{3}$\quad\\
\textbf{Quan Chen}$^{3}$\quad
\textbf{Peng Jiang}$^{3}$\quad
\textbf{Wenjun Wu}$^{2, 1}$\quad
\textbf{Liu Liu}$^{2,1*}$\\[0.5em]
$^1$Hangzhou International Innovation Institute, Beihang University\\
$^2$School of Artificial Intelligence, Beihang University\\
$^3$Kuaishou Technology
% \\
% $^4$School of Statistics and Data Science, Capital University of Economics and Business
}
\maketitle
\begingroup
\renewcommand\thefootnote{*}
\footnotetext{Corresponding author: \texttt{liuliubh@buaa.edu.cn}}
\endgroup
\begin{abstract}
\vspace{-1.0em}
\definecolor{abscolor}{HTML}{EAF4F7}
\begin{tcolorbox}[
    enhanced,
    breakable,
    colback=abscolor,
    colframe=abscolor,
    boxrule=0pt,
    arc=6pt,
    left=12pt, right=12pt, top=8pt, bottom=8pt,
    drop fuzzy shadow={black!45},
]
\itshape
\quad \quad Reinforcement learning (RL) has become a cornerstone for aligning and unlocking the reasoning capabilities of large-scale models. At its core, the training loop of GRPO and its variants alternates between \emph{rollout sampling} and \emph{policy update}: the policy first samples rollouts from its action space, and then updates its parameters according to the advantages computed over them. Unlike supervised learning, where each gradient step is anchored to an explicit ground-truth target, the optimal gradient direction for updating model parameters in this setting is not known a priori; the high-quality rollouts drawn during the sampling stage therefore act as the implicit "teacher" that guides every parameter update. However, mainstream RL algorithms such as GRPO adopt a simple sampling scheme that conditions all rollouts on the same original prompt. When a task lies beyond the policy model's current capability, this sampling scheme rarely yields a high-quality rollout, leaving the policy model without a meaningful gradient direction when updating its parameters, which causes training to stall. To address this issue, we propose \textbf{FBOS-RL}, a \emph{ Feedback-Driven Bi-Objective Synergistic Reinforcement Learning } framework. Specifically, we let the model perform Feedback-Guided Exploration Enhancement based on the feedback provided by the environment, and on top of this we design two mutually reinforcing training objectives: \emph{Exploitation-oriented Policy Alignment} (EPA) and \emph{Exploration-oriented Capability Cultivation} (ECC). Extensive experiments demonstrate that EPA and ECC can mutually reinforce each other, forming a positive flywheel effect that significantly improves both the training efficiency and the final performance ceiling of reinforcement learning. Specifically, under both an identical number of rollouts and the same number of training steps, FBOS-RL learns substantially faster than GRPO and feedback-based baselines and ultimately attains a higher performance ceiling, while exhibiting higher policy entropy and lower gradient norms throughout training.
\end{tcolorbox}
\end{abstract}
\section{Introduction}
\begin{figure}[H]
    \centering
    \vspace{-0.6em}
    \includegraphics[width=0.95\linewidth]{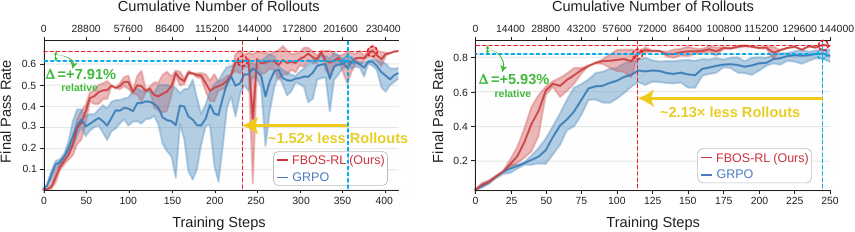}
    \vspace{-0.6em}
    \caption{\footnotesize
    Final pass rate on the TravelPlanner validation set during training: \textbf{FBOS-RL} (red) vs.\ vanilla GRPO (blue), on Llama-3.1-8B-Instruct (left) and Qwen3-14B (right). Bottom $x$-axis: training steps; top $x$-axis: cumulative rollouts; $y$-axis: final pass rate. The \emph{horizontal} dashed lines mark each method's best (peak) final pass rate (gap = relative gain $\Delta$); the \emph{vertical} dashed lines mark the cumulative rollouts at which each method first reaches GRPO's best (ratio = rollout speedup, up to $2.13\times$ fewer). Under both an identical total number of rollouts and the same number of training steps, FBOS-RL attains a higher ceiling \emph{and} reaches GRPO's best with far fewer rollouts.}
    \label{fig:teaser}
    \vspace{-0.4em}
\end{figure}
\vspace{0.4em}
Reinforcement learning (RL) plays a pivotal role in the alignment and reasoning fine-tuning of large language models (LLMs)~\citep{ouyang2022instructgpt,bai2022constitutional,rafailov2023dpo,guo2025deepseek,shao2024deepseekmath}. In essence, the training process of GRPO and its variants alternates between two stages, \emph{rollout sampling} and \emph{policy update}~\citep{schulman2017ppo,shao2024deepseekmath,tang2025rlexploration}: the model first samples in its action space to draw multiple rollouts, and then updates its policy parameters according to these rollouts and their advantages. Unlike supervised learning, where the model is directly updated toward an explicit ground truth, in this setting the model does not know the optimal gradient direction a priori when its parameters are updated. The high-quality rollouts that the model happens to draw during the sampling stage in fact play the role of a ``teacher'' that guides the parameter update. Only when the model samples high-quality rollouts can the parameter update obtain a better gradient direction, thereby driving the model's capability forward.

\begin{figure}[t]
    \centering
    \includegraphics[width=\linewidth]{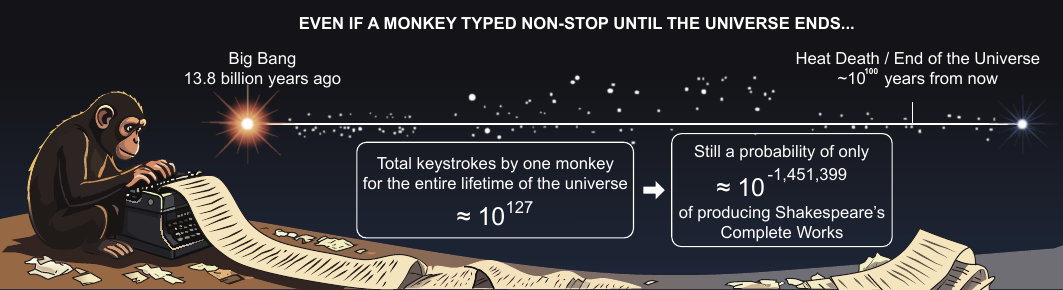}
    \caption{\small
    An illustrative analogy of the rollout-sampling stage in vanilla GRPO-style RL: a monkey randomly hitting keys on a typewriter is highly unlikely to ever produce the works of Shakespeare. Likewise, when a prompt exceeds the policy's current capability, simple sampling strategies rarely produce a high-quality rollout, leaving training without a meaningful gradient anchor.}
    \label{fig:monkey}
    \vspace{-0.2in}
\end{figure}

Despite the empirical success of mainstream RL algorithms in practice, prevailing large-scale-model RL algorithms (such as GRPO and its variants)~\citep{shao2024deepseekmath,yu2026dapo,guo2025deepseek} commonly adopt a single and blind sampling paradigm during the sampling stage: given the original prompt, the model directly generates multiple rollouts conditioned on it~\citep{shao2024deepseekmath,yu2026dapo}. This paradigm has a key limitation: when faced with complex tasks that exceed the model's current capability, the model tends to enter a regime of inefficient sampling, and rarely samples a high-quality rollout (just like a monkey banging on a typewriter, who could not produce the complete works of Shakespeare even before the universe ends, see Figure~\ref{fig:monkey}). In this situation, a single scalar reward can only tell the model that its current policy ``performs poorly'', but cannot indicate ``why'' it is poor or ``how to improve'' (akin to a student drilling problems without any explanation of past mistakes, whose progress is slow)~\citep{huang2023selfcorrect}. Because the model cannot obtain high-quality rollouts as optimization anchors during the sampling stage, it loses a reliable gradient direction when its parameters are updated, which eventually leads to low training efficiency and even prolonged training stagnation~\citep{yue2025doesrlincentivize,cui2025entropy,liu2025prorl}.

To address this issue, we propose \textbf{FBOS-RL}: \textbf{F}eedback-Driven \textbf{B}i-\textbf{O}bjective \textbf{S}ynergistic \textbf{R}einforcement \textbf{L}earning. The core innovation of FBOS-RL is that, through the guidance of feedback, the rollout-sampling stage is enhanced, and on top of this, two mutually promoting training objectives are designed, namely \emph{Exploitation-oriented Policy Alignment} (EPA) and \emph{Exploration-oriented Capability Cultivation} (ECC). By alternately optimizing these two objectives, FBOS-RL forms a positive bootstrapping flywheel, thereby substantially improving both the training efficiency and the final performance ceiling of reinforcement learning.
Although several recent works have attempted to introduce external feedback during the sampling stage, most of them merely treat feedback as a means of data augmentation, that is, they use feedback to generate higher-quality data and then let the model imitate it~\citep{madaan2023selfrefine,shinn2023reflexion,kumar2024scor,qu2024recursive}, while overlooking the cultivation of the model's own capability to sample better rollouts according to feedback. In contrast, FBOS-RL not only leverages feedback to enhance the quality of the rollouts obtained during the sampling stage, but also explicitly sets ``understanding feedback and correcting errors'' itself as an additional reinforcement learning objective. Through the alternating optimization of the two training objectives, a positive bootstrapping flywheel is formed, which improves both the training efficiency and the final performance ceiling. As illustrated in Figure~\ref{fig:teaser}, on the TravelPlanner benchmark FBOS-RL matches the best final pass rate of vanilla GRPO with roughly $1.52\times$ fewer rollouts on Llama-3.1-8B-Instruct and $2.13\times$ fewer rollouts on Qwen3-14B, demonstrating a substantially better training efficiency; meanwhile, under both an identical total number of rollouts and the same number of training steps, FBOS-RL further pushes the peak final pass rate beyond that of vanilla GRPO by a relative margin of $+7.91\%$ and $+5.93\%$ on the two models, respectively, evidencing a markedly higher final-performance ceiling. Beyond these main results, FBOS-RL also exhibits stronger exploration and better training stability: as shown in Figure~\ref{fig:actor_entropy} and Figure~\ref{fig:grad_norm}, FBOS-RL does not suffer from entropy collapse and consistently maintains a higher policy entropy than vanilla GRPO throughout training, while at the same time exhibiting a lower gradient norm.

% {\color{red}
Our contributions are summarized as follows:
% \begin{itemize}[topsep=4pt, itemsep=2pt, parsep=0pt, partopsep=0pt]
\begin{itemize}[leftmargin=*,itemsep=1pt]
        \item We point out the inefficiency and blindness of mainstream reinforcement learning algorithms (such as GRPO) during the rollout-sampling stage, and propose an interactive sampling mechanism, namely Feedback-Guided Exploration Enhancement, which effectively breaks the inefficient-sampling bottleneck on complex reasoning tasks.
    \item We propose the \textbf{Feedback-Driven Bi-Objective Synergistic Reinforcement Learning (FBOS-RL)} framework. By designing two mutually promoting training objectives (Exploitation-oriented Policy Alignment and Exploration-oriented Capability Cultivation), it forms a positive bootstrapping flywheel that substantially improves both the training efficiency and the final performance ceiling of reinforcement learning.
    \item Extensive experiments on different datasets and on models of different families and scales validate that FBOS-RL can substantially improve both the training efficiency and the final performance ceiling of reinforcement learning. Meanwhile, FBOS-RL also avoids entropy collapse, maintains higher policy entropy, and exhibits a lower gradient norm, indicating its stronger exploration capability and better training stability.
\end{itemize}
% }

% 做成表格：
% 	训练目标 1	训练目标 2
% 本质	利用 (Exploitation)	探索 (Exploration)
% 功能	让模型向采样到的高质量 rollout 靠拢	训练模型通过反馈探索到更优的 rollout

\section{Related works}
% \vspace{-0.15in}
\subsection{Reinforcement Learning for LLM Reasoning}
% \vspace{-0.1in}
Reinforcement learning (RL) has become a central tool for aligning LLMs and unlocking their reasoning capabilities~\citep{ouyang2022instructgpt,bai2022constitutional,rafailov2023dpo}. Building on PPO~\citep{schulman2017ppo}, scalable algorithms such as GRPO~\citep{shao2024deepseekmath} and DAPO~\citep{yu2026dapo} have driven remarkable progress on mathematical and code reasoning, exemplified by DeepSeek-R1~\citep{guo2025deepseek}. To improve credit assignment, process reward models~\citep{lightman2023letsverify,wang2024mathshepherd} and implicit process rewards~\citep{shi2025verifiers} provide step-level supervision, while ProRL~\citep{liu2025prorl} and entropy-aware optimization~\citep{cui2025entropy} extend the reasoning frontier through prolonged or stabilized training.
A growing body of work, however, points out that the rollout-sampling stage in mainstream RL pipelines is fundamentally blind: rollouts are sampled solely from the original prompt, and a scalar reward only signals that the policy ``performs poorly'' without indicating ``why'' or ``how to improve''~\citep{huang2023selfcorrect}. Recent studies confirm that vanilla RL struggles to push the policy beyond the base model's reasoning boundary~\citep{yue2025doesrlincentivize} and emphasize the under-investigated role of exploration~\citep{tang2025rlexploration,wang2025reinforcement}. Mitigations include reverse curricula~\citep{xi2024r3}, tree search guided RL~\citep{zheng2025grouprelativetreesearch}, learning from negative data~\citep{setlur2024rlnegative}, test-time compute scaling~\citep{snell2024scaling}, and self-play or self-rewarding schemes~\citep{chen2024selfplay,yuan2024selfrewarding}. Yet the gradient-driving rollouts are still produced by a single, feedback-free pass, leaving the core blindness largely unaddressed.

\subsection{Feedback-Driven Self-Refinement and Self-Correction}
% \vspace{-0.1in}

A complementary line of work investigates how natural-language feedback can help LLMs refine their outputs. Self-Refine~\citep{madaan2023selfrefine} and Reflexion~\citep{shinn2023reflexion} use self-critique or verbal reflection at inference time, and self-critiquing models~\citep{saunders2022selfcritiquing}, learning-to-self-correct~\citep{welleck2023selfcorrect}, and tool-augmented critiquing such as CRITIC~\citep{gou2024critic} further show that feedback-conditioned generation raises output quality. More recent works train the self-correction behavior itself, e.g., SCoRe~\citep{kumar2024scor} and recursive introspection~\citep{qu2024recursive}. Earlier studies caution that LLMs cannot reliably self-correct without external grounding~\citep{huang2023selfcorrect}, motivating verifiable signals~\citep{wang2024mathshepherd,lightman2023letsverify}. Off-policy guidance methods such as LUFFY~\citep{yan2025learning} additionally reweight the importance ratio to amplify learning on critical low-probability tokens.
Most of these approaches, however, treat feedback either as an inference-time scaffold or as data for imitation, and rarely close the loop between feedback-augmented exploration and the underlying policy optimization; the resulting prompt distribution shift is typically ignored, yielding biased gradients.
This fundamental limitation restricts the model’s ability to iteratively refine its behavior according to real-time feedback signals.

\section{Feedback-Driven Bi-Objective Synergistic Reinforcement Learning}
\label{sec:method}
\begin{figure}[t]
    \centering
    \includegraphics[width=\linewidth]{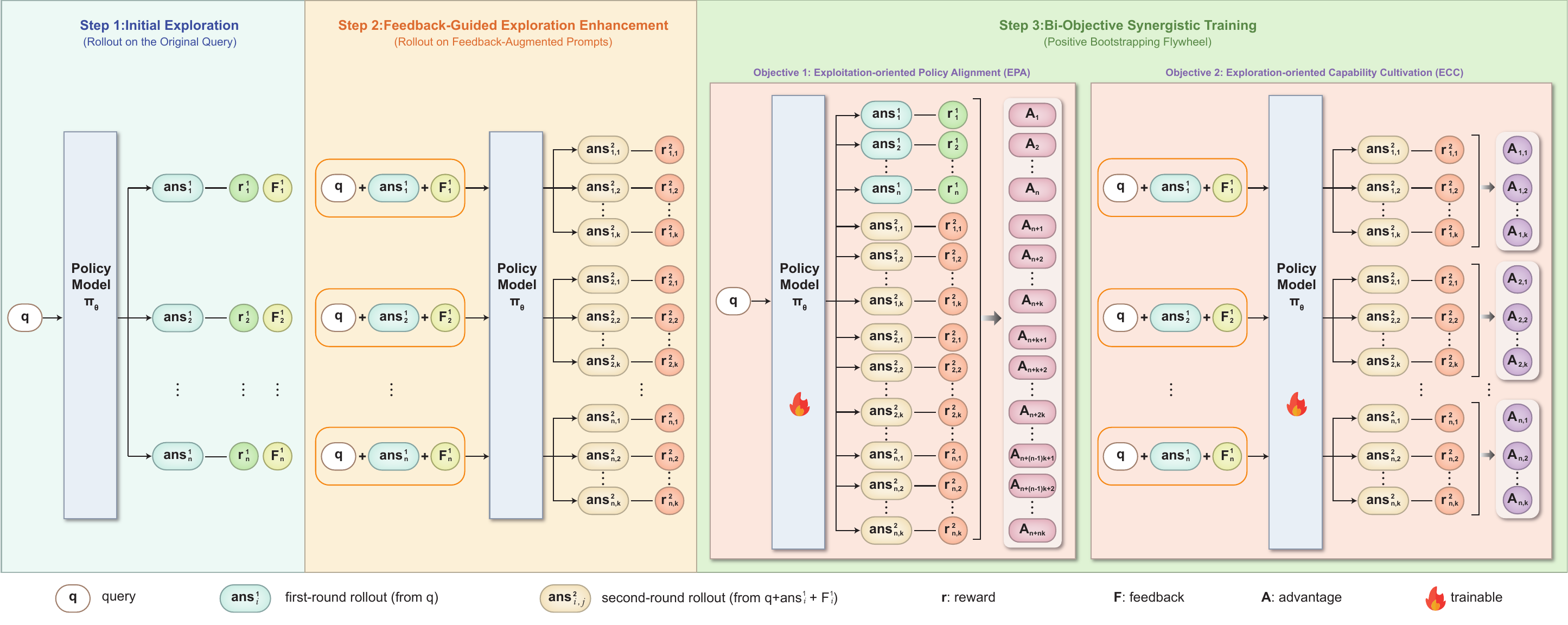}
    \caption{\small Overview of our Feedback-Driven Bi-Objective Synergistic RL (FBOS-RL) framework. In the sampling phase, the policy first generates $n$ initial rollouts from the original prompt $q$; a rule-based verifier produces a natural-language feedback for each, which is then concatenated with $q$ and the rollout to form a Feedback-Augmented Prompt (FAP) used for a second round of feedback-guided sampling. In the optimization phase, two complementary objectives are co-optimized: Exploitation-oriented Policy Alignment (EPA) via FA-GRPO and Exploration-oriented Capability Cultivation (ECC), which together induce a positive bootstrapping flywheel.}
    \label{fig:method_overview}
    \vspace{-0.2in}
\end{figure}

In this section, we introduce FBOS-RL, the Feedback-Driven Bi-Objective Synergistic Reinforcement Learning framework, whose overall pipeline is illustrated in Figure~\ref{fig:method_overview}. FBOS-RL performs reinforcement learning on a single policy model $\pi_\theta$ and does not rely on any external model. It substantially boosts both training efficiency and the attainable performance ceiling through two mutually reinforcing training objectives that together form a positive bootstrapping flywheel. Each training step of FBOS-RL is organized into three stages: the \emph{Initial Exploration} stage, the \emph{Feedback-Guided Exploration Enhancement} stage, and the \emph{Bi-Objective Synergistic Training} stage, which we elaborate on in turn below.

% \vspace{-0.15in}
\subsection{Initial Exploration}
% \vspace{-0.1in}

In the \emph{Initial Exploration} stage, corresponding to the leftmost ``Step 1: Initial Exploration'' block of Figure~\ref{fig:method_overview}, the policy model $\pi_\theta$ independently samples, in parallel, $n$ initial rollouts $\{\mathrm{ans}^1_i\}_{i=1}^{n} \sim \pi_\theta(\cdot \mid q)$ conditioned solely on the original prompt $q$. Each rollout $\mathrm{ans}^1_i$ with $i \in [n]$ is then evaluated by a rule-based verifier that returns both a scalar reward $r^1_i$ and a natural-language feedback $F^1_i$ that explicitly identifies the flaws in the current answer (e.g., a specific reasoning step that violates a constraint, an incorrect intermediate result, or a violation of the required output format). Unlike a single scalar reward, $F^1_i$ provides actionable, error-localized information that tells the model not only that $\mathrm{ans}^1_i$ is suboptimal, but also \textit{why} and \textit{where} it went wrong.

\subsection{Feedback-Guided Exploration Enhancement}
% \vspace{-0.1in}
\label{subsec:feedback_exploration}

To enhance the quality of the rollouts sampled during the sampling stage, so that the model can obtain higher-quality gradient guidance when its parameters are updated in the subsequent training phase, we proceed from the Initial Exploration stage to the \emph{Feedback-Guided Exploration Enhancement} stage, which corresponds to the orange block in Figure~\ref{fig:method_overview}. For each initial rollout $\mathrm{ans}^1_i$, we concatenate the original prompt $q$, the rollout itself, and its corresponding feedback $F^1_i$ to construct a \textit{Feedback-Augmented Prompt (FAP)}: $\tilde q_i \triangleq (q \oplus \mathrm{ans}^1_i \oplus F^1_i)$. The FAP is then used to re-prompt the policy model for a second round of sampling: for each $\tilde q_i$, the model samples $k$ rollouts $\{\mathrm{ans}^2_{i,j}\}_{j=1}^{k} \sim \pi_\theta(\cdot \mid \tilde q_i)$, and each $\mathrm{ans}^2_{i,j}$ is again scored by the same rule-based verifier to obtain a reward $r^2_{i,j}$.

In total, the two stages above produce $n + n\cdot k$ rollouts for each query, among which $n$ rollouts are generated by the model conditioned on the original prompt $q$, and the remaining $n \cdot k$ rollouts are generated conditioned on the feedback-augmented prompts $\tilde q_i$. The primary advantage of this mechanism lies in overcoming the blindness of the traditional sampling stage, and substantially boosting the quality of the rollouts sampled during the sampling stage. Instead of forcing the model to blindly sample rollouts from the original prompt $q$, which often yields low-quality rollouts when the task exceeds the model's current capability, we provide targeted feedback for each initial rollout. This feedback explicitly informs the model on \textit{how} to adjust its subsequent generation. Consequently, even when the model's initial capability is weak, the FAPs serve as a guiding scaffold, improving the quality of the second-round rollouts and enabling the model to align with the environment rapidly. This process is analogous to a student learning from a detailed explanation of their mistakes rather than blindly guessing the correct answer.

% \begin{figure}[t]
%     \centering
%     \includegraphics[width=0.5\linewidth]{Figures/student.pdf}
%     \caption{An intuitive analogy of our framework. \textbf{Left:} vanilla GRPO is akin to a student who can only see a numerical score and must blindly redo problems, making slow and inefficient progress. \textbf{Right:} our feedback-augmented exploration is akin to a student who receives detailed worked-out feedback for every mistake, enabling rapid and targeted improvement.}
%     \label{fig:student}
% \end{figure}

% \subsection{Bi-Objective Co-Evolutionary Training}
% \subsection{Bi-Objective Synergistic RL Training}
% \vspace{-0.14in}
\subsection{Bi-Objective Synergistic Training}
% \vspace{-0.1in}
\label{subsec:dual_objective}

Building upon the two sampling stages described above, we design two mutually reinforcing training objectives, namely \emph{Exploitation-oriented Policy Alignment} (EPA) and \emph{Exploration-oriented Capability Cultivation} (ECC), as illustrated in the rightmost green block of Figure~\ref{fig:method_overview}. At each training step, we sequentially update the policy parameters using these two objectives.

In this paper, the qualifiers ``exploitation-oriented'' and ``exploration-oriented'' carry operational meanings specific to our framework. By ``exploitation-oriented'' we mean exploiting the high-quality rollouts already collected as a learning signal for policy alignment. By ``exploration-oriented'' we mean cultivating the model's capability to sample better rollouts when conditioned on feedback-augmented prompts. These usages are scoped to GRPO-style training and are not intended to coincide with the classical action-selection trade-off in textbook reinforcement learning.

\vspace{0.5em}
\noindent\textbf{Objective 1: Exploitation-oriented Policy Alignment (EPA).}\\[0.25em]
We aggregate the $n$ initial rollouts $\{\mathrm{ans}^1_i\}_{i=1}^{n}$ (drawn from $\pi_{\theta_{\text{old}}}(\cdot\mid q)$) and the $n\cdot k$ rollouts $\{\mathrm{ans}^2_{i,j}\}_{i=1,j=1}^{n,k}$ produced in the Feedback-Guided Exploration Enhancement stage (drawn from $\pi_{\theta_{\text{old}}}(\cdot\mid \tilde q_i)$) into a single group
$
    \mathcal{G}_q \;\triangleq\; \{\mathrm{ans}^1_i\}_{i=1}^{n} \,\cup\, \{\mathrm{ans}^2_{i,j}\}_{i=1,j=1}^{n,\,k},
$
of size $N \triangleq n + n\cdot k$. Based on the rewards $\mathcal{R}_q\triangleq\{r^1_i\}_{i=1}^{n} \cup \{r^2_{i,j}\}_{i=1,j=1}^{n,k}$ associated with these rollouts, we compute a group-normalized advantage following GRPO~\citep{shao2024deepseekmath} for every rollout $a \in \mathcal{G}_q$ and reward $r(a)\in \mathcal{R}_q$ as
\begin{equation*}
    \hat A^{\text{EPA}}(a) \;=\; \frac{r(a) - \mu_{\mathcal{G}_q}}{\sigma_{\mathcal{G}_q} + \epsilon},
    \qquad
    \text{with}\quad
    \mu_{\mathcal{G}_q} = \frac{1}{N}\sum\nolimits_{a' \in \mathcal{G}_q} r(a'),\quad
    \sigma_{\mathcal{G}_q}^{2} = \frac{1}{N}\sum\nolimits_{a' \in \mathcal{G}_q} \big(r(a') - \mu_{\mathcal{G}_q}\big)^{2}.
    % \label{eq:advantage_epa}
\end{equation*}
For an initial rollout $\mathrm{ans}^1_i$ (drawn from $\pi_{\theta_{\text{old}}}(\cdot\mid q)$), the importance sampling ratio at token $t$ retains the standard form, i.e.,
\begin{equation*}
    \rho^{1}_{i,t}(\theta) \;=\; \frac{\pi_\theta(\mathrm{ans}^1_{i,t} \mid q,\, \mathrm{ans}^1_{i,<t})}{\pi_{\theta_{\text{old}}}(\mathrm{ans}^1_{i,t} \mid q,\, \mathrm{ans}^1_{i,<t})}.
\end{equation*}

However, for a rollout $\mathrm{ans}^2_{i,j}$ collected in the Feedback-Guided Exploration Enhancement stage, the behavior policy that actually drew it is $\pi_{\theta_{\text{old}}}(\cdot\mid \tilde q_i)$, whereas the policy we ultimately deploy at inference time only sees the original prompt $q$. By definition of an importance sampling ratio, the numerator must correspond to the target policy that we wish to optimize and deploy, while the denominator must correspond to the behavior policy that actually produced the sample. The corresponding token-level importance sampling ratio is therefore defined as
\begin{equation*}
    \rho^{2}_{i,j,t}(\theta) \;=\; \frac{\pi_\theta(\mathrm{ans}^2_{i,j,t} \mid q,\, \mathrm{ans}^2_{i,j,<t})}{\pi_{\theta_{\text{old}}}(\mathrm{ans}^2_{i,j,t} \mid \tilde q_i,\, \mathrm{ans}^2_{i,j,<t})}.
    % \label{eq:fa_grpo_ratio}
\end{equation*}
% This ratio asks ``how likely would $\pi_\theta$ have produced the refined answer \emph{without seeing the feedback}'' relative to ``how likely the behavior policy was to produce it \emph{with} the feedback'', so that the off-policy correction is aligned with the prompt distribution we actually want to optimize.

Inspired by Yan et al.~\citep{yan2025learning}, we additionally introduce a reweighting function $f(\rho) = \rho / (\rho + 0.1)$ that acts on $\rho^{2}_{i,j,t}(\theta)$, in order to strengthen the learning signal on low-probability tokens that the model has not yet mastered but that may correspond to critical reasoning steps.
We then treat $\mathcal{G}_q$ as a single group and perform GRPO training over it. Putting the above ingredients together, and applying the reweighting function $f(\cdot)$ exclusively to $\rho^{2}_{i,j,t}(\theta)$, the EPA objective is given by
\begin{equation}
    \mathcal{L}_{\text{EPA}}(\theta) \;=\; \mathcal{L}_{\text{EPA}}^{\text{init}}(\theta) \;+\; \mathcal{L}_{\text{EPA}}^{\text{FAP}}(\theta),
\label{eq:fa_grpo_loss}
\end{equation}
where $\mathcal{L}_{\text{EPA}}^{\text{init}}(\theta)$ accounts for the $n$ initial rollouts drawn from $\pi_{\theta_{\text{old}}}(\cdot\mid q)$ and $\mathcal{L}_{\text{EPA}}^{\text{FAP}}(\theta)$ accounts for the $n\cdot k$ rollouts drawn from $\pi_{\theta_{\text{old}}}(\cdot\mid \tilde q_i)$, respectively,
\begin{align}
\mathcal{L}_{\text{EPA}}^{\text{init}}(\theta) &= -\,\mathbb{E}
% _{[\,q\sim\mathcal{D},\; \mathrm{ans}^1_i \sim \pi_{\theta_{\text{old}}}(\cdot\mid q)\,]}
\Bigg[
\frac{1}{N}\sum_{i=1}^{n}\frac{1}{|\mathrm{ans}^1_i|}\sum_{t=1}^{|\mathrm{ans}^1_i|}
\min\!\Big(\rho^{1}_{i,t}(\theta)\,\hat A^{\text{EPA}}(\mathrm{ans}^1_i),\;
\mathrm{clip}_\text{init}\hat A^{\text{EPA}}(\mathrm{ans}^1_i)\Big) \Bigg],\nonumber
% \label{eq:fa_grpo_loss_init}
% \end{align},
% \begin{align}
% \small
% \begin{aligned}
\\
\mathcal{L}_{\text{EPA}}^{\text{FAP}}(\theta) &= -\,\mathbb{E}
% _{[\,q\sim\mathcal{D},\; \mathrm{ans}^1_i \sim \pi_{\theta_{\text{old}}}(\cdot\mid q),\; \mathrm{ans}^2_{i,j} \sim \pi_{\theta_{\text{old}}}(\cdot\mid \tilde q_i)\,]}
\Bigg[
\frac{1}{N}\sum_{i=1}^{n}\sum_{j=1}^{k}\frac{1}{|\mathrm{ans}^2_{i,j}|}\sum_{t=1}^{|\mathrm{ans}^2_{i,j}|}
\min\!\Big(f\!\big(\rho^{2}_{i,j,t}(\theta)\big)\,\hat A^{\text{EPA}}(\mathrm{ans}^2_{i,j}),\;
\mathrm{clip}_\text{FAP}\hat A^{\text{EPA}}(\mathrm{ans}^2_{i,j})\Big) \Bigg].\nonumber
% \end{aligned}
% \label{eq:fa_grpo_loss_fap}
\end{align}
Note that $\mathrm{clip}_\text{init}=\mathrm{clip}\!\big(\rho^{1}_{i,t}(\theta),\,1-\varepsilon,\,1+\varepsilon\big)\,$ and $\mathrm{clip}_\text{FAP}=\mathrm{clip}\!\big(f\!\big(\rho^{2}_{i,j,t}(\theta)\big),\,1-\varepsilon,\,1+\varepsilon\big)\,$.
The essence of EPA is therefore to pull the model's \emph{feedback-free} policy $\pi_\theta(\cdot\mid q)$ toward the high-quality rollouts discovered during the feedback-guided sampling stage, thereby achieving highly efficient exploitation of the collected high-quality rollouts. We empirically verify that EPA can be optimized in a stable manner: as shown in Appendices~\ref{app:epa_training_curves}, on both Llama-3.1-8B-Instruct and Qwen3-14B, the EPA training-set score steadily rises while its std steadily decreases, and the same trend holds across training subsets of different difficulty levels.

\vspace{0.5em}
\noindent\textbf{Objective 2: Exploration-oriented Capability Cultivation (ECC).}\\[0.25em]
% To train the model's ability to discover higher-quality rollouts guided by FAPs, we treat the $k$ rollouts generated in the second round for \textit{each} specific FAP as an independent group and perform standard GRPO training. This objective focuses exclusively on cultivating the model's \textbf{exploration} capability—specifically, 是根据包含反馈的FAPs探索到更优rollouts的能力.
To cultivate the model's ability to discover higher-quality rollouts when guided by FAPs, we treat the $k$ rollouts generated in the second round for \textit{each} specific FAP $\tilde q_i$ as an independent group $\mathcal{G}_{\tilde q_i} = \{\mathrm{ans}^2_{i,j}\}_{j=1}^{k}$ and perform standard GRPO training. This objective focuses exclusively on cultivating the model's \textbf{feedback-conditioned sampling capability}, i.e., its ability to discover better rollouts when conditioned on FAPs that contain feedback signals.

% 注意：这里需要给出优势 A 的数学表达式！
Based on the rewards $\{r^2_{i,j}\}_{j=1}^{k}$ associated with these rollouts, we compute a group-normalized advantage for each rollout $a \in \mathcal{G}_{\tilde q_i}$ following GRPO~\citep{shao2024deepseekmath} as
\begin{equation*}
    \hat A^{\text{ECC}}(a) \;=\; \frac{r(a) - \mu_{\tilde q_i}}{\sigma_{\tilde q_i} + \epsilon},
    \qquad
    \text{with}\quad
    \mu_{\tilde q_i} = \frac{1}{k}\!\!\sum\nolimits_{a' \in \mathcal{G}_{\tilde q_i}}\!\! r(a'),
    \quad
    \sigma_{\tilde q_i}^{2} = \frac{1}{k}\sum\nolimits_{a' \in \mathcal{G}_{\tilde q_i}} \big(r(a') - \mu_{\tilde q_i}\big)^{2}.
    \label{eq:advantage_ecc}
\end{equation*}
% $ \sum_{a}\ b$
% which contrasts each refined rollout against its siblings drawn from the same FAP, isolating the marginal value of \emph{how} the model leveraged the feedback.（which contrasts each ... 这句话需要调整。） 
Note that \(r(a)\) denotes the reward corresponding to the rollout \(a\). 
Because every rollout in $\mathcal{G}_{\tilde q_i}$ shares the same conditioning prompt $\tilde q_i$, the standard GRPO importance ratio applies without modification:
$
    \tilde\rho_{i,j,t}(\theta) \;=\; \frac{\pi_\theta(\mathrm{ans}^2_{i,j,t} \mid \tilde q_i,\, \mathrm{ans}^2_{i,j,<t})}{\pi_{\theta_{\text{old}}}(\mathrm{ans}^2_{i,j,t} \mid \tilde q_i,\, \mathrm{ans}^2_{i,j,<t})}.
$
The ECC objective is defined as follows:
% \begin{equation}
% \small
% \begin{aligned}
% &\mathcal{L}_{\text{ECC}}(\theta) \;=\; -\,\mathbb{E}_{[\,q\sim\mathcal{D},\; \mathrm{ans}^1_i \sim \pi_{\theta_{\text{old}}}(\cdot\mid q),\; \mathrm{ans}^2_{i,j} \sim \pi_{\theta_{\text{old}}}(\cdot\mid \tilde q_i)\,]}\Bigg[\\
% &\frac{1}{k}\sum_{j=1}^{k}\frac{1}{|\mathrm{ans}^2_{i,j}|}\sum_{t=1}^{|\mathrm{ans}^2_{i,j}|} \min\!\Big(\tilde\rho_{i,j,t}(\theta)\,\hat A^{\text{ECC}}(\mathrm{ans}^2_{i,j}),\;
% \mathrm{clip}\big(\tilde\rho_{i,j,t}(\theta),\,1-\varepsilon,\,1+\varepsilon\big)\,\hat A^{\text{ECC}}(\mathrm{ans}^2_{i,j})\Big) \Bigg].
% \end{aligned}
% \label{eq:ecc_loss}
% \end{equation}
\begin{align}
% \small
% \begin{aligned}
\mathcal{L}_{\text{ECC}}(\theta) 
= -\,\mathbb{E}\Bigg[
\frac{1}{k}\sum_{j=1}^{k}\frac{1}{|\mathrm{ans}^2_{i,j}|}\sum_{t=1}^{|\mathrm{ans}^2_{i,j}|} \min\!\Big(\tilde\rho_{i,j,t}(\theta)\,\hat A^{\text{ECC}}(\mathrm{ans}^2_{i,j}),\;
\mathrm{clip}_\text{ECC}\hat A^{\text{ECC}}(\mathrm{ans}^2_{i,j})\Big) \Bigg],\nonumber
% \end{aligned}
\label{eq:ecc_loss}
\end{align}
where $
% q\sim\mathcal{D},\; \mathrm{ans}^1_i \sim \pi_{\theta_{\text{old}}}(\cdot\mid q),\; \mathrm{ans}^2_{i,j} \sim \pi_{\theta_{\text{old}}}(\cdot\mid \tilde q_i),
\mathrm{clip}_\text{ECC}=\mathrm{clip}\big(\tilde\rho_{i,j,t}(\theta),\,1-\varepsilon,\,1+\varepsilon\big)\,$.
% By optimizing $\mathcal{L}_{\text{ECC}}$, the model explicitly learns to convert a flawed initial attempt and its critique into a substantially better rollout, turning the act of \emph{listening to feedback} into a first-class, learnable skill rather than a one-off prompting trick.
% ECC 是为了提升模型基于 FAPs 探索到更高质量 rollout 的能力，使其学会如何更好地根据反馈修正错误。
% （将这两个训练目标类比到学生做练习题）
% 这就像学生不断学习如何根据错题解析优化答案（对应ECC），并学习到：“奥，原来下次见到这个题目我应该这样道！”（对应EPA）。因此能更加高效地学习，进步神速。
By optimizing $\mathcal{L}_{\text{ECC}}$, ECC improves the model's ability to discover higher-quality rollouts based on FAPs, so that it learns how to better correct its mistakes according to the feedback. We empirically verify that ECC can be optimized in a stable manner as well: as shown in Appendices~\ref{app:ecc_training_curves}, on both Llama-3.1-8B-Instruct and Qwen3-14B, the ECC training-set score steadily rises while its std steadily decreases, and the same trend holds across training subsets of different difficulty levels.
The two objectives are closely analogous to how a student tackles practice problems: ECC corresponds to the act of refining answers based on the explanations of past mistakes, while EPA corresponds to gradually internalizing those lessons into the realization that ``ah, so next time I encounter this problem, I should solve it this way!'' Together, they enable the student to learn more efficiently and make rapid progress.

\subsection{The Positive Bootstrapping Flywheel Effect}
% \vspace{-0.1in}
\label{subsec:flywheel}

In what follows, we describe how EPA and ECC mutually reinforce each other and together give rise to a positive bootstrapping flywheel. The two directions of this mutual reinforcement are empirically validated by our ablation study in Section~\ref{app:ablation}.

% \begin{itemize}[leftmargin=*,itemsep=1pt]
% \vspace{-0.1in}
%     \item 
    % \textbf{ECC boosts EPA:}
    \textbf{ECC boosts EPA} (empirically verified in Section~\ref{app:ecc_boosts_epa})\textbf{:}
    ECC trains the model to discover higher-quality rollouts conditioned on FAPs, and these high-quality rollouts are subsequently injected into the group of EPA for GRPO training. Therefore, ECC raises the quality of the rollouts used for EPA training, thereby boosting EPA by providing it with a higher-quality gradient direction and elevating its policy-alignment ceiling.

    % \item 
    \textbf{EPA elevates ECC} (empirically verified in Section~\ref{app:epa_boosts_ecc})\textbf{:}
    At the very beginning of training, the model typically does not yet possess the capability to generate sufficiently good rollouts conditioned solely on the original prompt $q$, and frequently makes relatively low-level mistakes. Consequently, the feedbacks $\{F^1_i\}_{i=1}^{n}$ received by the rollouts $\{\mathrm{ans}^1_i\}_{i=1}^{n}$ produced in the Initial Exploration stage are mostly targeted at low-level errors (e.g., format errors). At this stage, ECC can only train the model's ability to discover better rollouts based on ``feedback targeting low-level errors'', and is hardly able to train its ability to discover better rollouts based on ``feedback targeting higher-level errors'' (since the model rarely receives feedback targeting higher-level errors at this point). EPA, in contrast, trains the model to generate better rollouts conditioned solely on the original prompt $q$. By bringing EPA into the loop, the model increasingly receives ``feedback targeting higher-level errors'' during the Initial Exploration stage, so that ECC can in turn train the model's ability to discover better rollouts based on ``feedback targeting higher-level errors'' (avoiding a mismatch between the model's capability and the difficulty of the feedback). The model begins to break through low-level errors, thereby triggering ``advanced feedback'' targeting deeper logical flaws. This capability progression enables ECC to continuously train the model's correction ability in an error space of ever-higher difficulty.
% \end{itemize}
\section{Experiments}
\subsection{Experimental Setup}
% \vspace{-0.1in}
\paragraph{Datasets.}
%
% 我们分别在 TravelPlanner 数据集\cite{xie2024travelplanner} 和 MiniF2F (Zheng et al., 2022) （Lean4 版本）上进行了实验。
% TravelPlanner 数据集是 xxx（简要介绍一下数据集）
% MiniF2F-Lean4 数据集是 xxx（简要介绍一下数据集）
% 对于TravelPlanner 数据集，我们将其训练集（45条）和验证集（180条）共计225条数据作为训练集；将其测试集（1000条）作为验证集。
% 对于MiniF2F-Lean4 数据集，我们将其验证集（244 题）作为训练集，将其测试集（244 题）作为验证集。
%
We conduct experiments on the TravelPlanner dataset~\cite{xie2024travelplanner} and the MiniF2F dataset~\cite{zheng2022minif2f} (Lean4 version), respectively.
% {\color{red}The TravelPlanner dataset is xxx (a brief introduction to the dataset).
% The MiniF2F-Lean4 dataset is xxx (a brief introduction to the dataset).}
The TravelPlanner dataset evaluates the ability of LLMs to perform complex, multi-constraint planning in real-world scenarios, while MiniF2F-Lean4 benchmarks their capacity for formal mathematical reasoning and proof generation within the Lean 4 formalization environment. Detailed descriptions of both datasets are provided in Appendix~\ref{app:dataset_details}.
For the TravelPlanner dataset, we use its training set (45 samples) and validation set (180 samples), totaling 225 samples, as our training set, and adopt its test set (1000 samples) as our validation set.
For the MiniF2F-Lean4 dataset, we use its validation set (244 samples) as our training set, and use its test set (244 samples) as our validation set.

% \vspace{-0.1in}
\paragraph{Models}
% We train both the Qwen-3-14B model and the Llama-3.1-8B-Instruct model.

% 在 TravelPlanner 数据集上，我们分别对 Llama-3.1-8B-Instruct 模型 和 Qwen3-14B 模型进行了实验，在 MiniF2F-Lean4 数据集上，我们对 Qwen3.5-9B 模型进行了实验。

On the TravelPlanner dataset, we conduct experiments on the Llama-3.1-8B-Instruct model and the Qwen3-14B model, respectively. On the MiniF2F-Lean4 dataset, we conduct experiments on the Qwen3.5-27B model.
% 所有训练均在 NVIDIA H200 GPU 上进行。
All training is conducted on NVIDIA H200 GPUs.

% \vspace{-0.1in}
\paragraph{Baseline Methods.}
We evaluate our method against GRPO~\cite{shao2024deepseekmath}, as well as three comparison methods designed by ourselves: FBOS-RL w/o EPA, FBOS-RL w/o ECC, and GRPO w/ Extra Update. Specifically:
\begin{itemize}[leftmargin=*]
    % \item \noindent\textbf{GRPO}~\cite{shao2024deepseekmath}: A standard group-relative policy optimization baseline that estimates the advantage of each rollout by comparing it against the average reward of the rollout group sharing the same prompt, without an additional value network.
    \item \noindent\textbf{FBOS-RL w/o EPA}: A variant of FBOS-RL that removes the EPA training objective, which is used to ablate the contribution of the EPA objective.
    \item \noindent\textbf{FBOS-RL w/o ECC}: A variant of FBOS-RL that removes the ECC training objective, which is used to ablate the contribution of the ECC objective.
    \item \noindent\textbf{GRPO w/ Extra Update}: A variant of GRPO that adds one extra update step at each training step on top of GRPO, which is used to control for the effect of additional gradient updates introduced by FBOS-RL. This baseline rules out the possibility that the gains of FBOS-RL stem merely from the additional parameter update introduced by ECC; even under this strengthened baseline, FBOS-RL still substantially outperforms it, confirming that the improvements come from the synergistic flywheel rather than from the extra update itself. Detailed results of this controlled experiment on both the Qwen3-14B and Llama-3.1-8B-Instruct models are provided in Section~\ref{app:extra_update}.
\end{itemize}
\vspace{-0.1in}
To ensure a fair comparison, we keep the order of the training data identical across runs and ensure that the number of rollouts per training step is strictly the same. To mitigate the inherent stochasticity of reinforcement learning, we independently repeat each experiment three times.
In addition, on the TravelPlanner dataset, we also include the following baselines:

\begin{itemize}[leftmargin=*]
\vspace{-0.1in}
\item \noindent\textbf{Greedy Search}
To evaluate the performance of traditional search algorithms in TravelPlanner, we adopt the greedy search strategy as one of the baselines. Greedy Search focuses on cost minimization as its core objective. Among transportation options, it selects the one with the lowest cost; for dining, it chooses restaurants with the lowest average expenditure; for accommodation, it selects the cheapest option; and for sightseeing, it arranges attractions by randomly selecting them each day. For a 5-day or 7-day travel plan, select the top 1 to 2 cities as destinations from the returned city search results.
\item  \noindent\textbf{Sole-Planning Mode}
We focus on the sole-planning mode of the TravelPlan task. In this mode, the model is
provided in advance with sufficient and necessary reference information required for reasoning and planning. This setting is used to evaluate the model’s ability to perform complex reasoning and planning directly based on the given information. The baselines under the sole-planning mode include the following models and strategies:
\begin{itemize}[leftmargin=0.35cm]
    \item \textit{Models:} 
    GPT-3.5-Turbo, GPT-4-Turbo, GPT-4o (version: 2024-11-20), Mixtral-8×7B-MoE, Gemini Pro, Qwen3-8B-Instruct, DeepSeek-R1.
    \item \textit{Strategies}: Our baselines include the following strategies: Direct, CoT, ReAct, Reflexion, and prompt reflect. Specifically, "Direct" refers to prompting the model to directly generate the final travel plan; "CoT" refers to prompting the model to reason step by step before producing the final travel plan; "ReAct" refers to prompting the model to solve the task by alternating between Thought, Action, and Observation steps; "Reflexion" refers to prompting the model to perform self-reflection before generating the final travel plan; "Prompt reflect" refers to prompting the model to reflect on its previous reasoning before producing the final travel plan.
\end{itemize}

\end{itemize}
\vspace{-0.1in}

\providecolor{sigmaBG}{HTML}{D9FFDD}
\begin{table*}[t]
\caption{\small Experimental results comparing FBOS-RL with different baselines on TravelPlanner with five criteria}
\makebox[\textwidth][c]{
\centering
\small
\setlength{\tabcolsep}{3.5pt}
\renewcommand{\arraystretch}{0.95}
\begin{tabular}{@{\hspace{0pt}}>{\raggedright\arraybackslash}p{3.8cm}ccccc@{}}
\toprule
\multirow{2}{*}{\textbf{Methods}}
    & \multicolumn{2}{c}{\textbf{Commonsense Constraint}}
        & \multicolumn{2}{c}{\textbf{Hard Constraint}}
            & \multirow{2}{*}{\textbf{Final Pass Rate}} \\
 \cmidrule(lr){2-3} \cmidrule(lr){4-5}
    & Micro & Macro
        & Micro & Macro
            & \\
\midrule
$\text{Greedy Search}$           &72.0 &0 &52.4 &31.8 &0 \\
$\text{Mixtral-8x7B-MoE}_\text{Direct}$   &67.0 &3.7 &3.9 &1.6 &0.7 \\
$\text{GPT-3.5-Turbo}_\text{Direct}$           &59.5 &2.7 &9.5 &4.4 &0.6 \\
$\text{GPT-3.5-Turbo}_\text{CoT} $          &64.4 &2.3 &9.8 &3.8 &0.4 \\
$\text{GPT-3.5-Turbo}_\text{ReAct}$           &45.9 &2.5 &10.7 &3.1 &0.7 \\
$\text{GPT-3.5-Turbo}_\text{Reflexion}$           &52.1 &2.2 &9.9 &3.8 &0.6 \\
$\text{GPT-4-Turbo}_\text{Direct}$           &80.6 &15.2 &44.3 &23.1 &4.4 \\
$\text{GPT-4o-2024-11-20}_\text{CoT}$           &84.18 &25.9 &49.87 &26.6 &7 \\
$\text{Gemini Pro}_\text{Direct}$           &64.7 &7.9 &10.6 &4.7 &2.1 \\
$\text{DeepSeek-R1}_\text{CoT}$ &92.65 &55.2 &74.15 &62 &40 \\
$\text{Qwen3-8B}_\text{CoT}$           &72.04 &10.7 &28 &21.8 &5.9 \\
$\text{Qwen3-8B}_\text{prompt reflect}$           &73.84 &14.6 &27.95 &24 &8.3 \\
\hline
$\text{Llama-3.1-8B-Instruct}$ &57.12 &1.13 &6.75 &3.53 &0.43 \\
$\text{Llama-3.1-8B-Instruct}_\text{GRPO}$ &\textbf{98.83} &91.4 &\textbf{84.54} &69.7 &65.2 \\
\rowcolor{sigmaBG}
\textbf{$\text{Llama-3.1-8B-Instruct}_\text{FBOS-RL}$} &98.81 &\textbf{94} &84.11 &\textbf{71.7} &\textbf{68.7} \\
\hline
$\text{Qwen3-14B}$ &66.62 &6.13 &19.03 &15.5 &2.9 \\
$\text{Qwen3-14B}_\text{GRPO}$ &97.83 &89.5 &94.28 &91.4 &84.2 \\
\rowcolor{sigmaBG}
\textbf{$\text{Qwen3-14B}_\text{FBOS-RL}$} &\textbf{98.5} &\textbf{91.8} &\textbf{96.07} &\textbf{92.9} &\textbf{87.4} \\
\bottomrule
\end{tabular}
}
\label{tab:main-results}
\end{table*}

\noindent\textbf{Evaluation Metrics.}   We adopt standard metrics~\cite{xie2024travelplanner} to evaluate the performance of our method, including Commonsense Constraint Pass Rate, Hard Constraint Pass Rate, and Final Pass Rate. Detailed mathematical definitions of all the above metrics are provided in Appendix~\ref{app:eval_metrics}.
% The Commonsense Constraint Pass Rate covers eight commonsense dimensions (e.g., whether the visited cities are reasonable, whether restaurants/attractions are non-repetitive, whether all daily activities take place in the correct city) and evaluates whether the model can incorporate commonsense knowledge into the generated plan without being explicitly instructed. The Hard Constraint Pass Rate measures whether the generated plan satisfies the hard constraints explicitly given in the query (e.g., dietary, accommodation, transportation, and budget constraints). The Final Pass Rate denotes the proportion of plans that simultaneously satisfy all commonsense and hard constraints among all evaluated plans. For both Commonsense and Hard Constraint Pass Rate, we adopt two evaluation strategies, micro and macro, which respectively measure the model's ability to follow individual constraints and the full set of constraints. 
% 对于 MiniF2F-Lean4 数据集中的每条数据，我们采用如下评估方式：
% 如果模型给出了正确的证明，则给 +1 分；
% 如果模型给出了格式正确的回答，但证明写错了或者有步骤没证明完，则给 0 分；
% 如果模型给出的回答的格式有误，则给 -1 分。
% 我们对每条数据都采取上述评估方式，最终对分数取平均，作为 xxx。
For each sample in the MiniF2F-Lean4 dataset, we adopt the following evaluation scheme:
if the model gives a correct proof, it is awarded $+1$ point;
if the model gives a response in the correct format but the proof is wrong or some steps are not finished, it is awarded $0$ point;
if the model's response is in an incorrect format, it is awarded $-1$ point.
We apply the above evaluation scheme to every sample, and finally average the scores as the average score.

\subsection{Main Results}
% \vspace{-0.1in}

\begin{figure}[htbp]
  \centering
  \includegraphics[width=1\textwidth]{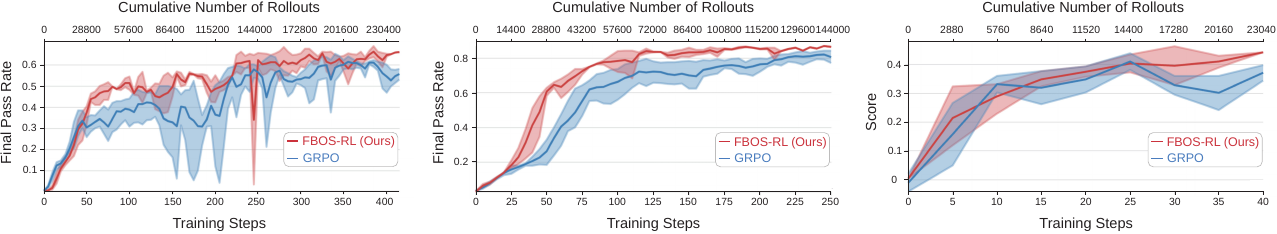}
  \vspace{-0.2in}
  \caption{
  \small
   Performance on the validation set during training: our method (FBOS-RL) vs.\ vanilla GRPO. Left: final pass rate of the Llama-3.1-8B-Instruct model on TravelPlanner. Middle: final pass rate of the Qwen3-14B model on TravelPlanner. Right: validation score of the Qwen3.5-27B model on MiniF2F-Lean4. The bottom $x$-axis denotes the number of training steps, the top $x$-axis denotes the cumulative number of rollouts, and the $y$-axis denotes the final pass rate (for TravelPlanner) or the validation score (for MiniF2F-Lean4) on the validation set.
  }
  \label{fig:main_results_comparison_curve}
\end{figure}

\rsubhead{Training Efficiency and Final Performance Ceiling.}
We assess FBOS-RL along two axes: \emph{training efficiency}, i.e., how quickly the model reaches a given level of validation performance, as measured by the number of training steps and cumulative rollouts; and the \emph{final performance ceiling}, i.e., the best validation performance ultimately attained. Table~\ref{tab:main-results} reports the best validation performance on TravelPlanner against an extensive collection of baselines, covering representative prompting-based strategies on a wide range of proprietary and open-source LLMs as well as GRPO trained on the same backbones; FBOS-RL attains the highest \emph{Final Pass Rate} under both the Llama-3.1-8B-Instruct backbone (\textbf{68.7}) and the Qwen3-14B backbone (\textbf{87.4}), confirming a clearly higher performance ceiling than every baseline considered. Since the table only reports the peak validation numbers and does not characterize the trajectory along which that ceiling is reached, we further plot the training-time validation curves of FBOS-RL versus vanilla GRPO in Figure~\ref{fig:main_results_comparison_curve}: on TravelPlanner with the Llama-3.1-8B-Instruct model (left) and the Qwen3-14B model (middle), and on MiniF2F-Lean4 with the Qwen3.5-27B model (right). The bottom $x$-axis denotes the number of training steps, the top $x$-axis denotes the cumulative number of rollouts, and the $y$-axis denotes the final pass rate (for TravelPlanner) or the validation score (for MiniF2F-Lean4) on the validation set. In terms of \emph{training efficiency}, FBOS-RL matches the best final pass rate of vanilla GRPO with roughly $1.52\times$ fewer rollouts on Llama-3.1-8B-Instruct and $2.13\times$ fewer rollouts on Qwen3-14B. In terms of the \emph{final performance ceiling}, under an identical total number of rollouts and an identical number of training steps, FBOS-RL further pushes the peak final pass rate beyond that of vanilla GRPO by a relative margin of $+7.91\%$ on Llama-3.1-8B-Instruct and $+5.93\%$ on Qwen3-14B.
% \begin{figure}[H]
%   \centering
%   \begin{subfigure}[t]{0.49\textwidth}
%     \centering
%     \includegraphics[width=\textwidth]{Figures/llama_3_1_8b/val-core:travelplanner:final_pass_rate_v1.pdf}
%     \caption{Llama-3.1-8B-Instruct.}
%     \label{fig:llama_3_1_8b/val-core:travelplanner:final_pass_rate_v1}
%   \end{subfigure}
%   \hfill
%   \begin{subfigure}[t]{0.49\textwidth}
%     \centering
%     \includegraphics[width=\textwidth]{Figures/qwen3_14b/val-core:travelplanner:final_pass_rate_v1.pdf}
%     \caption{Qwen3-14B.}
%     \label{fig:qwen3_14b/val-core:travelplanner:final_pass_rate_v1}
%   \end{subfigure}
%   \caption{
%   \small
%     Final pass rate on validation set.
%   }
%   \label{fig:val-core:travelplanner:final_pass_rate_v1}
%   % \vspace{-0.25in}
% \end{figure}
% 从图中可以看出，在 TravelPlanner 数据集上，在 Llama-3.1-8B-Instruct 模型 和 Qwen-3-14B 模型上，我们的方法相比于普通GRPO方法分别提升了 xxx 个点。
% 这里要算p值。p 值为 0.036
% TravelPlanner 数据集对数据的难度进行了划分（划分为"easy", "medium", "hard"）
% 我们还分别统计了训练过程中，模型在验证集中不同难度 level （分为"easy", "medium", "hard"）的数据上的 final pass rate。
\rsubhead{Fine-grained Analysis: Difficulty Levels and Constraint Types.}
To understand more precisely \emph{where} the gains of FBOS-RL over vanilla GRPO come from, we further break down the validation performance along two axes: by sample \emph{difficulty} (which probes how the gap behaves as the planning problem becomes harder) and by \emph{constraint type} (which probes whether the gains are driven by commonsense reasoning, by satisfaction of explicit hard constraints, or by both).

\begin{figure}[htbp]
  \centering
  \includegraphics[width=1\textwidth]{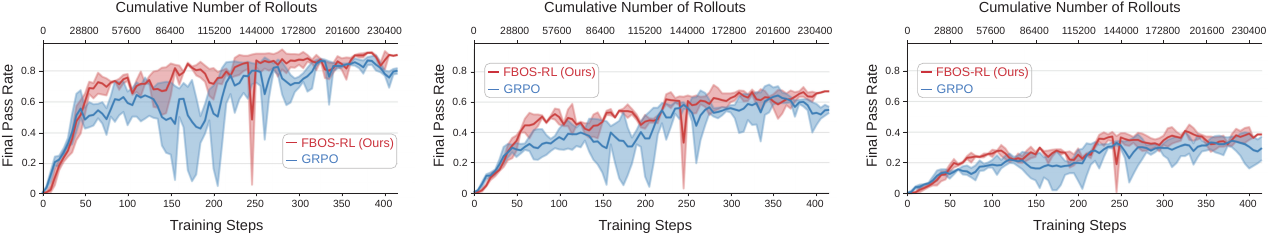}
  \vspace{-0.2in}
  \caption{
  \small
   Final pass rate of the Llama-3.1-8B-Instruct model on the TravelPlanner validation set during training, broken down by difficulty level: easy (left), medium (middle), and hard (right). Our method (FBOS-RL) is compared with vanilla GRPO.
  }
  \label{fig:llama_final_pass_rate_difficulty}
\end{figure}

\begin{figure}[htbp]
  \centering
  \includegraphics[width=1\textwidth]{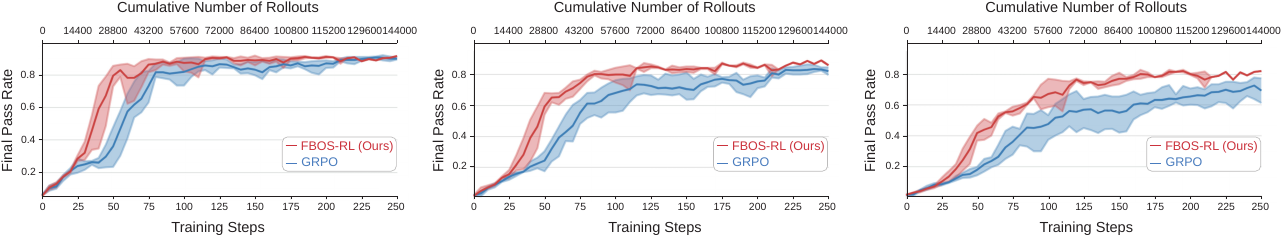}
  \vspace{-0.2in}
  \caption{
  \small
   Final pass rate of the Qwen3-14B model on the TravelPlanner validation set during training, broken down by difficulty level: easy (left), medium (middle), and hard (right). Our method (FBOS-RL) is compared with vanilla GRPO.
  }
  \label{fig:qwen3_14b_final_pass_rate_difficulty}
\end{figure}

\noindent\textit{Breakdown by difficulty.} The TravelPlanner dataset partitions samples by difficulty into ``easy'', ``medium'', and ``hard''. We track the final pass rate on each difficulty subset of the validation set throughout training. The results for the Llama-3.1-8B-Instruct model and the Qwen3-14B model are reported in Figure~\ref{fig:llama_final_pass_rate_difficulty} and Figure~\ref{fig:qwen3_14b_final_pass_rate_difficulty}, respectively. Across both backbones, FBOS-RL outperforms vanilla GRPO at every difficulty level; importantly, the margin of improvement is not uniform but widens noticeably as the difficulty increases. This pattern is especially pronounced on Qwen3-14B, indicating that the advantage of FBOS-RL grows with task difficulty. It is precisely on the harder, multi-constraint instances, where vanilla GRPO struggles to discover and reinforce high-quality rollouts, that FBOS-RL provides the largest gains.

\begin{figure}[htbp]
  \centering
  \includegraphics[width=1\textwidth]{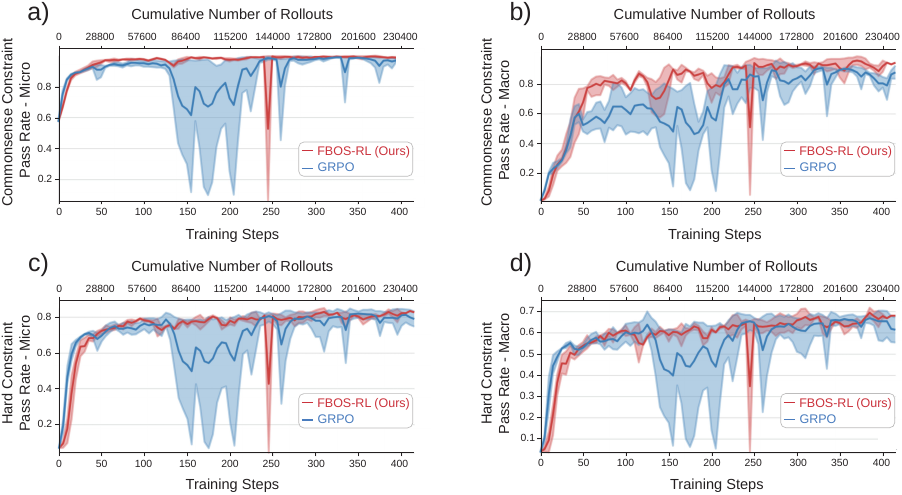}
  \vspace{-0.2in}
  \caption{
  \small
   Commonsense and hard constraint pass rates of the Llama-3.1-8B-Instruct model on the TravelPlanner validation set during training: Commonsense Constraint Pass Rate (Micro) (a), Commonsense Constraint Pass Rate (Macro) (b), Hard Constraint Pass Rate (Micro) (c), and Hard Constraint Pass Rate (Macro) (d). Our method (FBOS-RL) is compared with vanilla GRPO.
  }
  \label{fig:llama_constraint_pass_rate}
\end{figure}

\begin{figure}[htbp]
  \centering
  \includegraphics[width=1\textwidth]{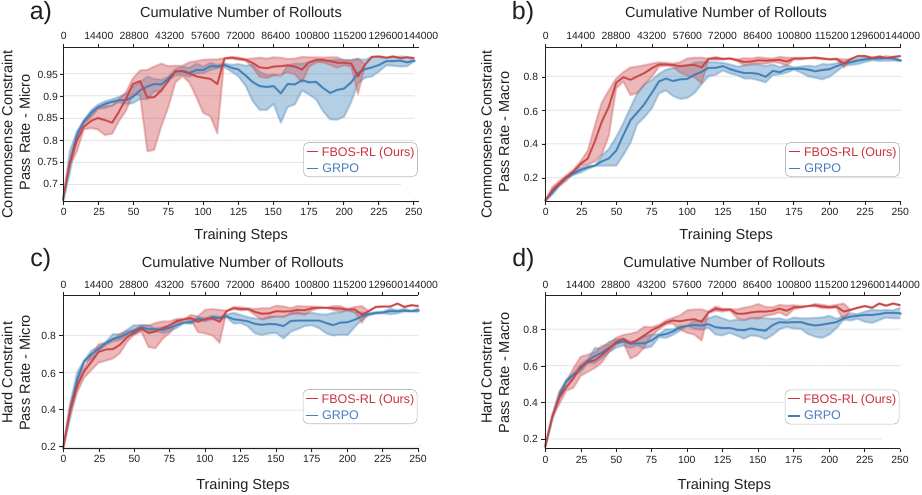}
  \vspace{-0.2in}
  \caption{
  \small
   Commonsense and hard constraint pass rates of the Qwen3-14B model on the TravelPlanner validation set during training: Commonsense Constraint Pass Rate (Micro) (a), Commonsense Constraint Pass Rate (Macro) (b), Hard Constraint Pass Rate (Micro) (c), and Hard Constraint Pass Rate (Macro) (d). Our method (FBOS-RL) is compared with vanilla GRPO.
  }
  \label{fig:qwen3_14b_constraint_pass_rate}
\end{figure}

\noindent\textit{Breakdown by constraint type.} By definition (see Appendix~\ref{app:eval_metrics}), the Final Pass Rate is computed at the plan level and only credits a plan when \emph{all} commonsense and \emph{all} hard constraints are simultaneously satisfied, which limits its resolution in two respects: (i) it operates at the plan level rather than the individual-constraint level, so it cannot reflect improvements measured at the per-constraint granularity captured by the \emph{Micro} statistics; and (ii) it aggregates commonsense and hard constraints into a single ``all satisfied'' indicator, so it cannot reveal how the two constraint families behave separately. To obtain a finer-grained picture along both axes, we additionally track the Commonsense Constraint Pass Rate (both Micro and Macro) and the Hard Constraint Pass Rate (both Micro and Macro) on the validation set throughout training; detailed mathematical definitions of these metrics are provided in Appendix~\ref{app:eval_metrics}. The results for the Llama-3.1-8B-Instruct model and the Qwen3-14B model are shown in Figure~\ref{fig:llama_constraint_pass_rate} and Figure~\ref{fig:qwen3_14b_constraint_pass_rate}, respectively. FBOS-RL surpasses vanilla GRPO on all four metrics for both backbones, indicating that the improvements of FBOS-RL come from both stronger commonsense reasoning and better satisfaction of explicit hard constraints, rather than from a single source.
\rsubhead{Stronger Exploration and Better Training Stability.}
Beyond the validation accuracy itself, we further measure the actor policy entropy and the gradient norm during training.

\begin{figure}[htbp]
  \centering
  \includegraphics[width=1\textwidth]{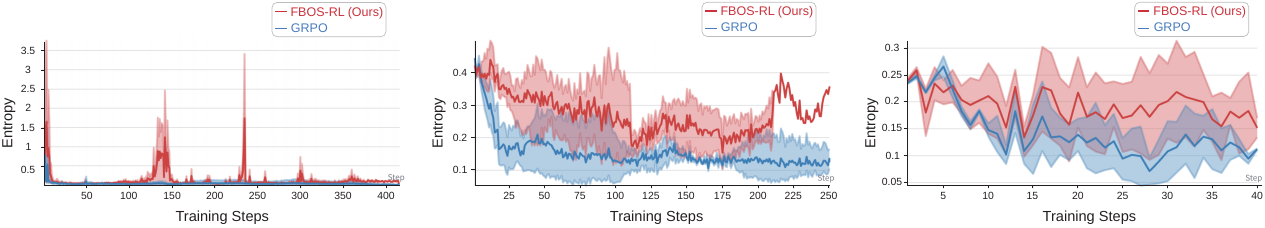}
  \vspace{-0.1in}
  \caption{
  \small
   Actor entropy during training: our method (FBOS-RL) vs.\ vanilla GRPO. Left: Llama-3.1-8B-Instruct on TravelPlanner. Middle: Qwen3-14B on TravelPlanner. Right: Qwen3.5-27B on MiniF2F-Lean4. Across all three settings, our method does not suffer from entropy collapse and consistently maintains higher entropy than vanilla GRPO.
  }
  \label{fig:actor_entropy}
\end{figure}

\noindent\textit{Stronger exploration.} GRPO can be prone to the entropy collapse phenomenon~\cite{cui2025entropy,yu2026dapo,wang2025arbitrary}, in which the actor's output distribution rapidly contracts onto a narrow set of high-probability actions during training, causing the policy to become overly deterministic and to lose the capacity to discover new, potentially higher-quality rollouts. As shown in Figure~\ref{fig:actor_entropy}, across all three settings (Llama-3.1-8B-Instruct and Qwen3-14B on TravelPlanner, and Qwen3.5-27B on MiniF2F-Lean4), the actor entropy of FBOS-RL remains consistently higher than that of vanilla GRPO and exhibits no sign of entropy collapse, indicating that FBOS-RL preserves a broader exploration distribution than vanilla GRPO throughout training.

\begin{figure}[htbp]
  \centering
  \includegraphics[width=1\textwidth]{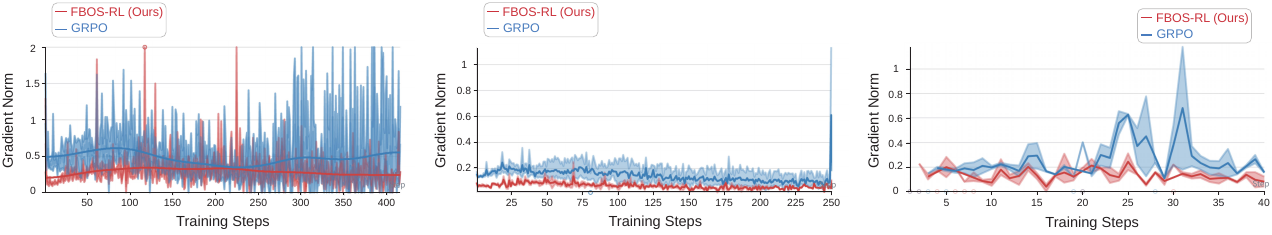}
  \vspace{-0.1in}
  \caption{
  \small
   Gradient norm during training: our method (FBOS-RL) vs.\ vanilla GRPO. Left: Llama-3.1-8B-Instruct on TravelPlanner. Middle: Qwen3-14B on TravelPlanner. Right: Qwen3.5-27B on MiniF2F-Lean4. Across all three settings, our method exhibits a lower gradient norm than vanilla GRPO, indicating better training stability.
  }
  \label{fig:grad_norm}
\end{figure}

% \noindent\textit{Better training stability.} A natural concern is that the stronger exploration could come at the cost of noisier or more unstable updates. The gradient-norm trajectories in Figure~\ref{fig:grad_norm} show that this is not the case: across all three settings, FBOS-RL exhibits a visibly lower and smoother gradient norm than vanilla GRPO throughout training, indicating that its parameter updates are better-conditioned and less prone to the abrupt spikes that often plague RL fine-tuning. This stability is consistent with the synergistic flywheel between EPA and ECC: ECC supplies EPA with higher-quality, feedback-corrected rollouts, which in turn yields smaller advantage variance and hence less noisy policy-gradient estimates. Taken together, FBOS-RL therefore expands the exploration budget \emph{and} tightens the optimization trajectory at the same time, two properties that are usually in tension under vanilla GRPO.
\noindent\textit{Better training stability.} We further examine the gradient norm during training. As shown in Figure~\ref{fig:grad_norm}, across all three settings, FBOS-RL exhibits a lower gradient norm than vanilla GRPO for the vast majority of training steps, indicating more stable optimization dynamics under the same training settings.

\begin{figure}[htbp]
  \centering
  \includegraphics[width=0.5\textwidth]{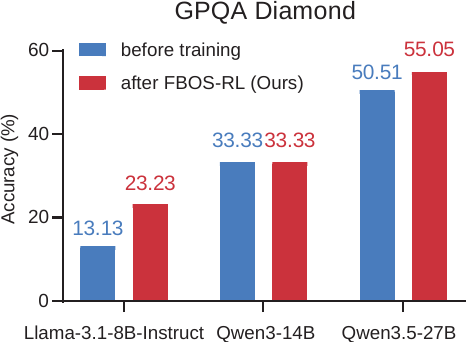}
  \caption{
  \small
   Out-of-distribution generalization to the GPQA-Diamond dataset: comparison of three models (Llama-3.1-8B-Instruct, Qwen3-14B, and Qwen3.5-27B) before training and after training with our FBOS-RL method. All models use greedy decoding during inference. The Llama-3.1-8B-Instruct and Qwen3-14B models are trained with FBOS-RL on the TravelPlanner dataset, while the Qwen3.5-27B model is trained with FBOS-RL on the MiniF2F-Lean4 dataset. None of the models are further trained on GPQA-Diamond, so this evaluation directly measures the out-of-distribution generalization gains brought by FBOS-RL.
  }
  \label{fig:ood}
\end{figure}

\rsubhead{Out-of-Distribution Generalization.}
% We further evaluate the trained models on the GPQA-Diamond dataset, with results shown in Figure~\ref{fig:ood}. All models use greedy decoding during inference. The Llama-3.1-8B-Instruct and Qwen3-14B models are trained only on TravelPlanner with FBOS-RL, while the Qwen3.5-27B model is trained only on MiniF2F-Lean4 with FBOS-RL; none of these models are trained on GPQA-Diamond. Compared with the untrained Llama-3.1-8B-Instruct model, the Llama-3.1-8B-Instruct model trained with FBOS-RL on TravelPlanner improves by 10.1 percentage points on GPQA-Diamond, corresponding to a relative improvement of 76.92\%. For Qwen3-14B, training with FBOS-RL on TravelPlanner does not lead to any performance degradation on GPQA-Diamond. For Qwen3.5-27B, training with FBOS-RL on MiniF2F-Lean4 improves GPQA-Diamond accuracy by 4.54 percentage points over the untrained model, corresponding to a relative improvement of 8.99\%.
We further evaluate the GPQA-Diamond performance of each model before training and after FBOS-RL training, with results shown in Figure~\ref{fig:ood}. All models use greedy decoding during inference. The Llama-3.1-8B-Instruct and Qwen3-14B models are trained only on TravelPlanner with FBOS-RL, while the Qwen3.5-27B model is trained only on MiniF2F-Lean4 with FBOS-RL; none of these models are trained on GPQA-Diamond. Compared with the untrained Llama-3.1-8B-Instruct model, the Llama-3.1-8B-Instruct model trained with FBOS-RL on TravelPlanner improves by 10.1 percentage points on GPQA-Diamond, corresponding to a relative improvement of 76.92\%. For Qwen3-14B, training with FBOS-RL on TravelPlanner does not lead to any performance degradation on GPQA-Diamond. For Qwen3.5-27B, training with FBOS-RL on MiniF2F-Lean4 improves GPQA-Diamond accuracy by 4.54 percentage points over the untrained Qwen3.5-27B model, corresponding to a relative improvement of 8.99\%.

% \subsection{Ablation Study}
\subsection{Ablation Study of Mutual Reinforcement between EPA and ECC}
\label{app:ablation}

To demonstrate that training Objective 1 (Exploitation-oriented Policy Alignment, EPA) and Objective 2 (Exploration-oriented Capability Cultivation, ECC) can mutually reinforce each other, forming a positive flywheel (bootstrapping) effect, we conduct in-depth experiments on the Qwen3-14B model.

\subsubsection{Objective 2 (ECC) Boosts Objective 1 (EPA)}
\label{app:ecc_boosts_epa}

We design a baseline that only optimizes Objective 1 (EPA) during training.

Objective 2 (ECC) trains the model to sample higher-quality rollouts conditioned on the Feedback-Augmented Prompt (FAP). These high-quality rollouts are then injected into the GRPO group used by Objective 1 (EPA). In this way, Objective 2 (ECC) effectively raises the quality of the rollouts that Objective 1 (EPA) leverages during the sampling stage, thereby boosting Objective 1 (EPA) (and significantly raising the ceiling of policy alignment).
(ECC, which cultivates the feedback-conditioned sampling capability, provides EPA, which exploits the already-sampled high-quality rollouts, with high-quality gradient guidance: it teaches the model how to produce better rollouts based on feedback.)

\begin{figure}[htbp]
  \centering
  \includegraphics[width=1\textwidth]{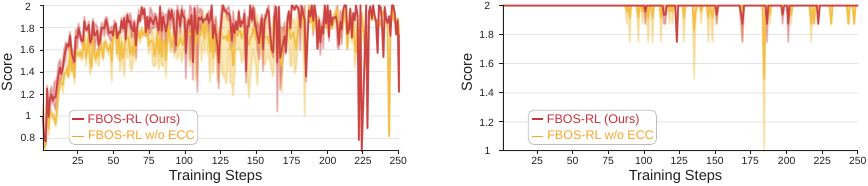}
  \vspace{-0.2in}
  \caption{
  \small
   Mean quality (left) and max quality (right) of rollouts generated by the Qwen3-14B model conditioned on the Feedback-Augmented Prompt (FAP) at each training step on the training set: our method vs.\ the baseline that only trains Objective 1 (EPA).
  }
  \label{fig:ecc_boosts_epa:fap_mean_max}
\end{figure}

Figure~\ref{fig:ecc_boosts_epa:fap_mean_max} reports, at each training step on the training set, the average quality and the best (max) quality of rollouts generated by the model conditioned on the Feedback-Augmented Prompt (FAP).

\begin{figure}[htbp]
  \centering
  \includegraphics[width=1\textwidth]{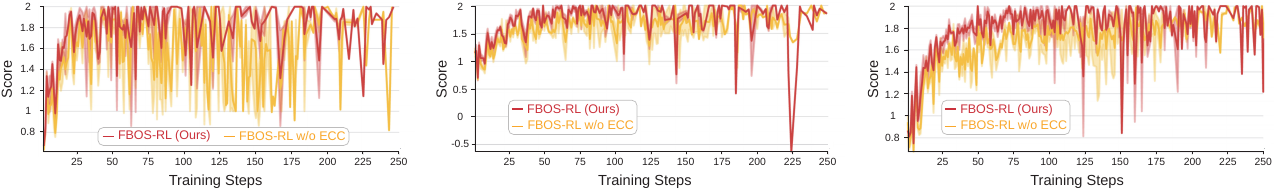}
  \vspace{-0.2in}
  \caption{
  \small
   Mean quality of FAP-conditioned rollouts on the training set, broken down by difficulty level: easy (left), medium (middle), and hard (right). Our method vs.\ the baseline that only trains Objective 1 (EPA).
  }
  \label{fig:ecc_boosts_epa:fap_mean_by_difficulty}
\end{figure}

\begin{figure}[htbp]
  \centering
  \includegraphics[width=1\textwidth]{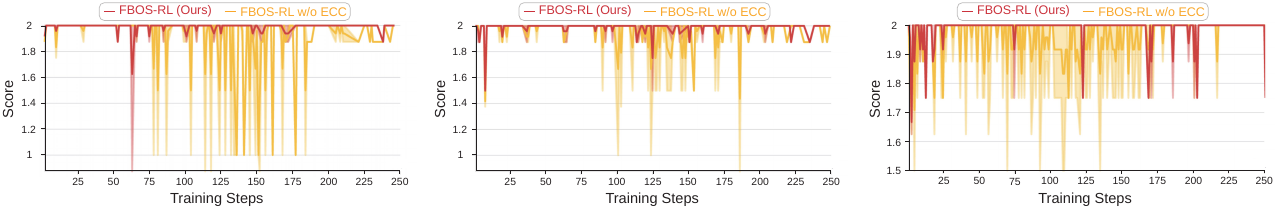}
  \vspace{-0.2in}
  \caption{
  \small
   Max quality of FAP-conditioned rollouts on the training set, broken down by difficulty level: easy (left), medium (middle), and hard (right). Our method vs.\ the baseline that only trains Objective 1 (EPA).
  }
  \label{fig:ecc_boosts_epa:fap_max_by_difficulty}
\end{figure}

We further analyze FAP-conditioned rollouts by difficulty level in Figure~\ref{fig:ecc_boosts_epa:fap_mean_by_difficulty} and Figure~\ref{fig:ecc_boosts_epa:fap_max_by_difficulty}. Figure~\ref{fig:ecc_boosts_epa:fap_mean_by_difficulty} reports the mean quality of rollouts generated by the model conditioned on the Feedback-Augmented Prompt (FAP), broken down by difficulty level. Figure~\ref{fig:ecc_boosts_epa:fap_max_by_difficulty} reports the max quality of rollouts generated by the model conditioned on the Feedback-Augmented Prompt (FAP), broken down by difficulty level.

Figures~\ref{fig:ecc_boosts_epa:fap_mean_max}, \ref{fig:ecc_boosts_epa:fap_mean_by_difficulty}, and~\ref{fig:ecc_boosts_epa:fap_max_by_difficulty} show that introducing Objective 2 (ECC) leads the model to generate higher-quality rollouts under the FAP, with both mean and max quality improved across every difficulty level.

\begin{figure}[htbp]
  \centering
  \includegraphics[width=0.6\textwidth]{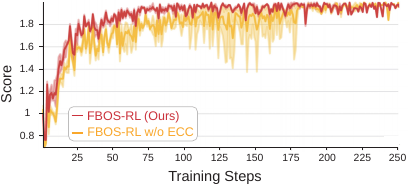}
  % \vspace{-0.2in}
  \caption{
  \small
   Mean quality of rollouts generated during the entire sampling phase (initial sampling and FAP-guided second-round sampling combined) at each training step on the training set: our method vs.\ the baseline that only trains Objective 1 (EPA).
  }
  \label{fig:ecc_boosts_epa:total_mean}
\end{figure}

Figure~\ref{fig:ecc_boosts_epa:total_mean} reports, at each training step on the training set, the average quality of rollouts generated during the sampling phase, including both the initial sampling and the second-round FAP-guided sampling.

\begin{figure}[htbp]
  \centering
  \includegraphics[width=1\textwidth]{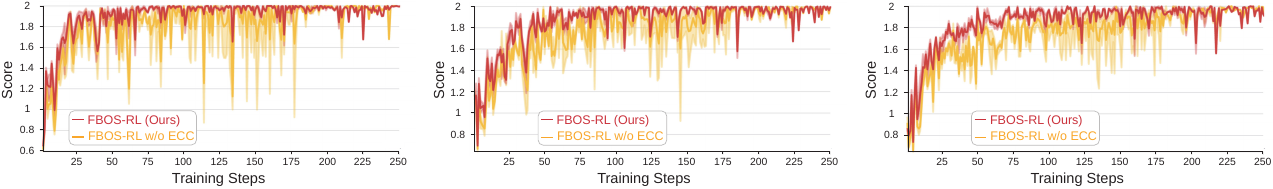}
  \vspace{-0.2in}
  \caption{
  \small
   Mean quality of rollouts generated during the sampling phase on the training set, broken down by difficulty level: easy (left), medium (middle), and hard (right). Our method vs.\ the baseline that only trains Objective 1 (EPA).
  }
  \label{fig:ecc_boosts_epa:total_mean_by_difficulty}
\end{figure}

\begin{figure}[htbp]
  \centering
  \includegraphics[width=1\textwidth]{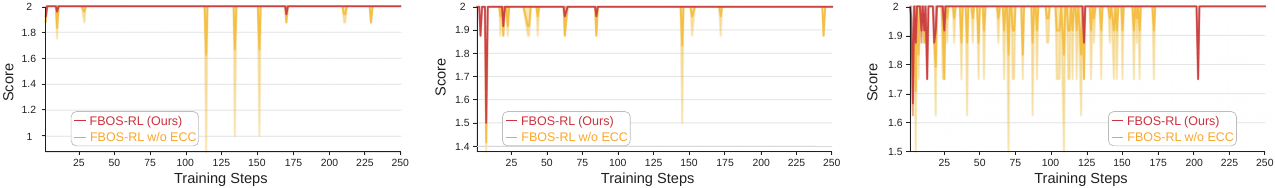}
  \vspace{-0.2in}
  \caption{
  \small
   Max quality of rollouts generated during the sampling phase on the training set, broken down by difficulty level: easy (left), medium (middle), and hard (right). Our method vs.\ the baseline that only trains Objective 1 (EPA).
  }
  \label{fig:ecc_boosts_epa:total_max_by_difficulty}
\end{figure}

We further report the mean and max quality of sampling-phase rollouts for each difficulty level in Figure~\ref{fig:ecc_boosts_epa:total_mean_by_difficulty} and Figure~\ref{fig:ecc_boosts_epa:total_max_by_difficulty}. Figure~\ref{fig:ecc_boosts_epa:total_mean_by_difficulty} reports the mean quality of sampling-phase rollouts at each difficulty level. Figure~\ref{fig:ecc_boosts_epa:total_max_by_difficulty} reports the max quality of sampling-phase rollouts at each difficulty level.

Figures~\ref{fig:ecc_boosts_epa:total_mean}, \ref{fig:ecc_boosts_epa:total_mean_by_difficulty}, and~\ref{fig:ecc_boosts_epa:total_max_by_difficulty} show that introducing Objective 2 (ECC) significantly improves the quality of rollouts discovered by the model during the sampling phase.

\begin{figure}[htbp]
  \centering
  \includegraphics[width=0.6\textwidth]{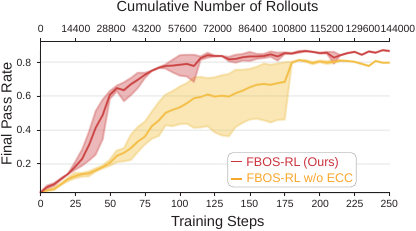}
  % \vspace{-0.2in}
  \caption{
  \small
   Final pass rate on the TravelPlanner validation set: our method vs.\ the baseline that only trains Objective 1 (EPA).
  }
  \label{fig:ecc_boosts_epa:final_pass_rate}
\end{figure}

\begin{figure}[htbp]
  \centering
  \includegraphics[width=1\textwidth]{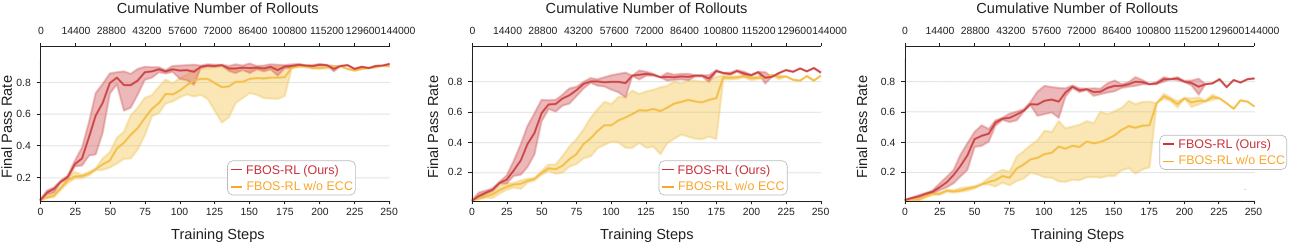}
  \vspace{-0.2in}
  \caption{
  \small
   Final pass rate on the TravelPlanner validation set across different difficulty levels: easy (left), medium (middle), and hard (right). Our method vs.\ the baseline that only trains Objective 1 (EPA).
  }
  \label{fig:ecc_boosts_epa:final_pass_rate_by_difficulty}
\end{figure}

\begin{figure}[htbp]
  \centering
  \includegraphics[width=1\textwidth]{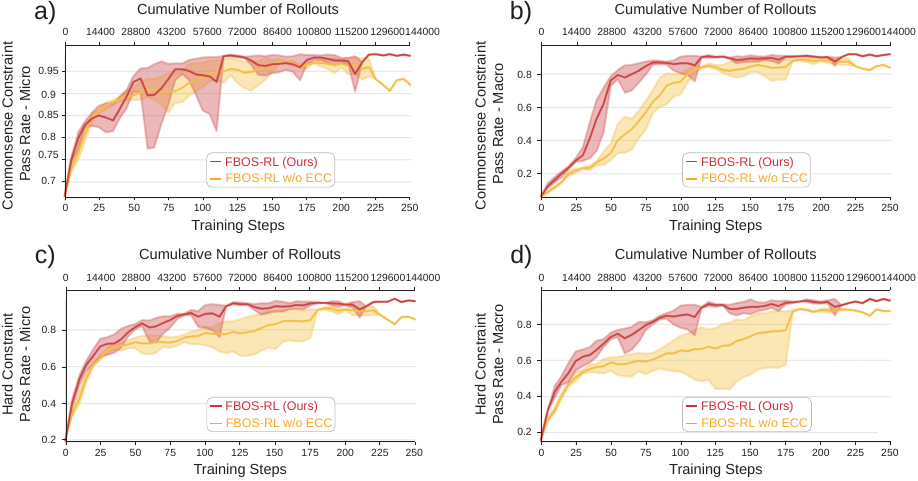}
  \vspace{-0.2in}
  \caption{
  \small
   Commonsense and hard constraint pass rates (micro and macro) on the TravelPlanner validation set: our method vs.\ the baseline that only trains Objective 1 (EPA).
  }
  \label{fig:ecc_boosts_epa:constraint_pass_rate}
\end{figure}

Figure~\ref{fig:ecc_boosts_epa:final_pass_rate} shows that our method significantly outperforms the baseline on the validation set. Figure~\ref{fig:ecc_boosts_epa:final_pass_rate_by_difficulty} reports, during training, the final pass rate on the validation set for each difficulty level (``easy'', ``medium'', ``hard''). As difficulty increases, the lead of our method grows larger. Figure~\ref{fig:ecc_boosts_epa:constraint_pass_rate} reports the following four metrics on the validation set during training: Commonsense Constraint Pass Rate (Micro), Commonsense Constraint Pass Rate (Macro), Hard Constraint Pass Rate (Micro), and Hard Constraint Pass Rate (Macro). Our method significantly outperforms the baseline on all four metrics.

These results demonstrate that Objective 2 (ECC) can effectively boost Objective 1 (EPA).

\subsubsection{Objective 1 (EPA) Boosts Objective 2 (ECC)}
\label{app:epa_boosts_ecc}

We design a baseline that only optimizes Objective 2 (ECC) during training.

\begin{figure}[htbp]
  \centering
  \includegraphics[width=0.6\textwidth]{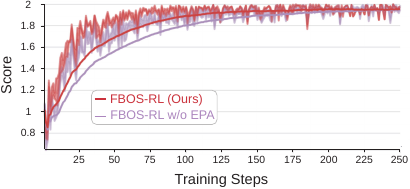}
  % \vspace{-0.2in}
  \caption{
  \small
   Mean quality of rollouts generated during the entire sampling phase at each training step on the training set: our method vs.\ the baseline that only trains Objective 2 (ECC).
  }
  \label{fig:epa_boosts_ecc:total_mean}
\end{figure}

\begin{figure}[htbp]
  \centering
  \includegraphics[width=1\textwidth]{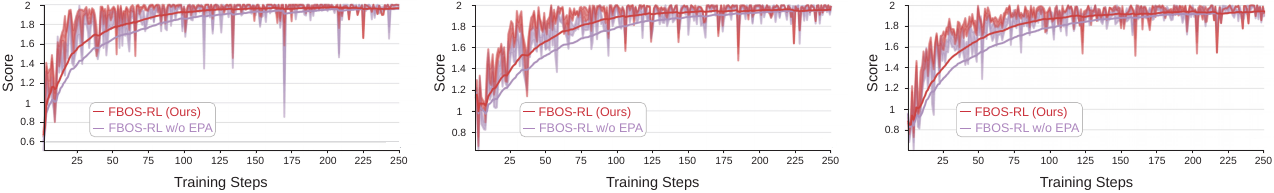}
  \vspace{-0.2in}
  \caption{
  \small
   Mean quality of rollouts generated during the sampling phase on the training set, broken down by difficulty level: easy (left), medium (middle), and hard (right). Our method vs.\ the baseline that only trains Objective 2 (ECC).
  }
  \label{fig:epa_boosts_ecc:total_mean_by_difficulty}
\end{figure}

Figure~\ref{fig:epa_boosts_ecc:total_mean} reports, at each training step on the training set, the average quality of rollouts generated by the model during the sampling phase, including both the initial sampling and the second-round sampling guided by the Feedback-Augmented Prompt (FAP). Figure~\ref{fig:epa_boosts_ecc:total_mean_by_difficulty} further reports the mean quality of sampling-phase rollouts at each difficulty level.

Figures~\ref{fig:epa_boosts_ecc:total_mean} and~\ref{fig:epa_boosts_ecc:total_mean_by_difficulty} show that introducing Objective 1 (EPA) significantly improves the quality of rollouts discovered by the model during the sampling phase.

\begin{figure}[htbp]
  \centering
  \includegraphics[width=0.6\textwidth]{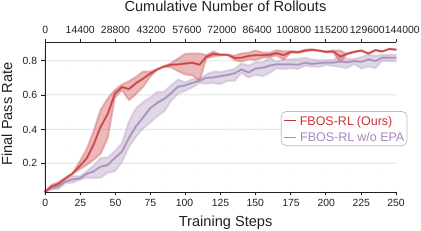}
  % \vspace{-0.2in}
  \caption{
  \small
   Final pass rate on the TravelPlanner validation set: our method vs.\ the baseline that only trains Objective 2 (ECC).
  }
  \label{fig:epa_boosts_ecc:final_pass_rate}
\end{figure}

\begin{figure}[htbp]
  \centering
  \includegraphics[width=1\textwidth]{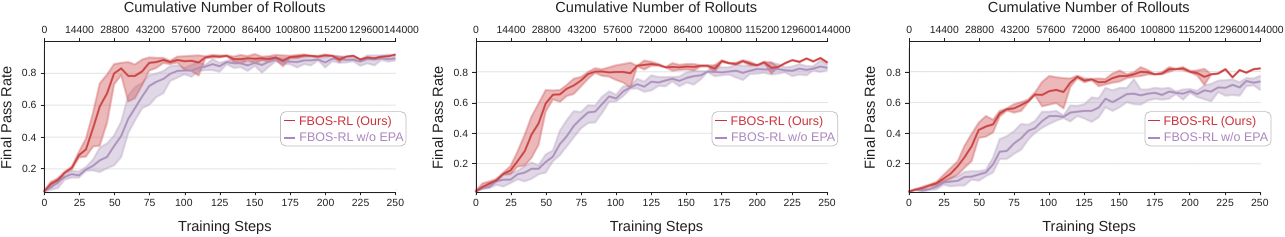}
  \vspace{-0.2in}
  \caption{
  \small
   Final pass rate on the TravelPlanner validation set across different difficulty levels: easy (left), medium (middle), and hard (right). Our method vs.\ the baseline that only trains Objective 2 (ECC).
  }
  \label{fig:epa_boosts_ecc:final_pass_rate_by_difficulty}
\end{figure}

\begin{figure}[htbp]
  \centering
  \includegraphics[width=1\textwidth]{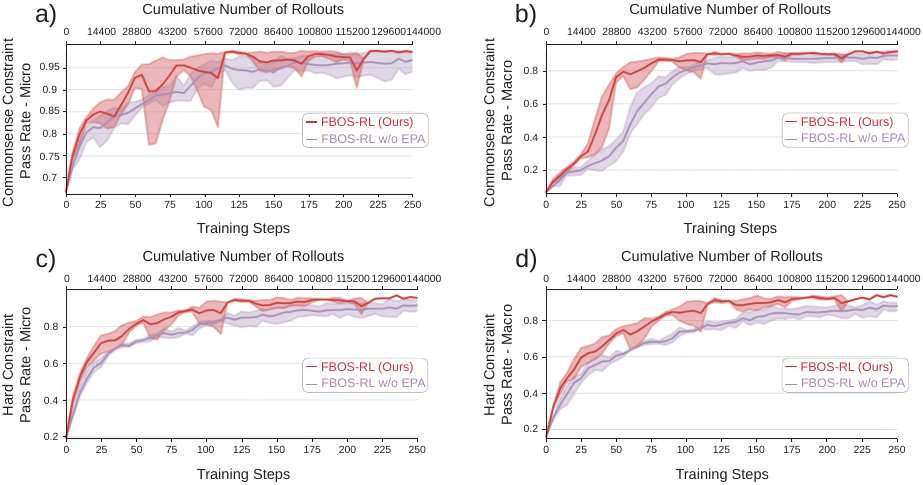}
  \vspace{-0.2in}
  \caption{
  \small
   Commonsense and hard constraint pass rates (micro and macro) on the TravelPlanner validation set: our method vs.\ the baseline that only trains Objective 2 (ECC).
  }
  \label{fig:epa_boosts_ecc:constraint_pass_rate}
\end{figure}

Figure~\ref{fig:epa_boosts_ecc:final_pass_rate} shows that our method significantly outperforms the baseline on the validation set. Figure~\ref{fig:epa_boosts_ecc:final_pass_rate_by_difficulty} reports, during training, the final pass rate on the validation set for each difficulty level (``easy'', ``medium'', ``hard''). Figure~\ref{fig:epa_boosts_ecc:constraint_pass_rate} reports the following four metrics on the validation set during training: Commonsense Constraint Pass Rate (Micro), Commonsense Constraint Pass Rate (Macro), Hard Constraint Pass Rate (Micro), and Hard Constraint Pass Rate (Macro). Our method significantly outperforms the baseline on all four metrics.

These results demonstrate that Objective 1 (EPA) can effectively boost Objective 2 (ECC).

\subsection{Controlling for the Number of Parameter Updates}
\label{app:extra_update}

Since our method performs two parameter updates per training step, while standard GRPO performs only one parameter update per training step, we design the following experiment to control for the effect of the number of parameter updates.

In our method, at each training step, Objective 1 (EPA) first performs one parameter update using all 72 rollouts, after which Objective 2 (ECC) performs an additional parameter update using the 64 rollouts produced by the second-round sampling.
We construct a strengthened baseline (referred to as \textit{GRPO w/ Extra Update}): at each training step, standard GRPO is also given one additional parameter update. Concretely, it first performs one parameter update using all 72 sampled rollouts, and then randomly samples 64 rollouts from these 72 to perform an additional parameter update, thereby matching the number of parameter updates in our method.

We conduct this controlled experiment on both the Qwen3-14B model (Section~\ref{app:extra_update_qwen}) and the Llama-3.1-8B-Instruct model (Section~\ref{app:extra_update_llama}). The two settings share an identical experimental protocol and only differ in the underlying base model.

\subsubsection{Qwen3-14B}
\label{app:extra_update_qwen}

\begin{figure}[htbp]
  \centering
  \includegraphics[width=0.6\textwidth]{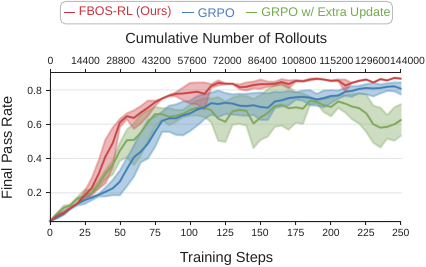}
  % \vspace{-0.2in}
  \caption{
  \small
   Final pass rate of the Qwen3-14B model on the TravelPlanner validation set: our method vs.\ the GRPO w/ Extra Update baseline.
  }
  \label{fig:extra_update:final_pass_rate}
\end{figure}

\begin{figure}[htbp]
  \centering
  \includegraphics[width=1\textwidth]{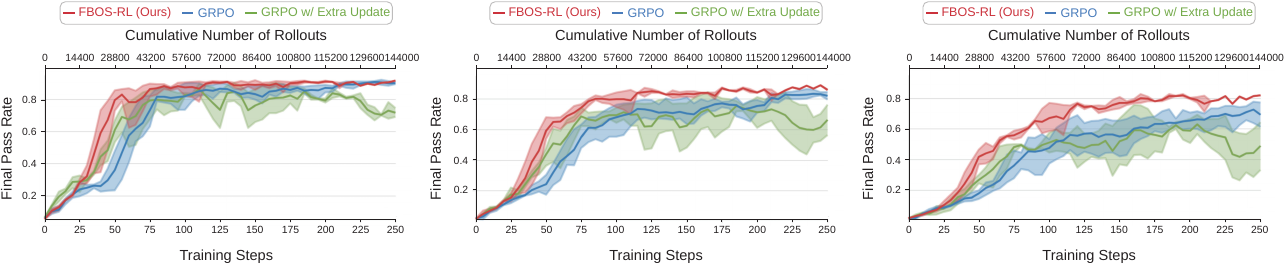}
  \vspace{-0.2in}
  \caption{
  \small
   Final pass rate of the Qwen3-14B model on the TravelPlanner validation set across different difficulty levels: easy (left), medium (middle), and hard (right). Our method vs.\ the GRPO w/ Extra Update baseline.
  }
  \label{fig:extra_update:final_pass_rate_by_difficulty}
\end{figure}

\begin{figure}[htbp]
  \centering
  \includegraphics[width=1\textwidth]{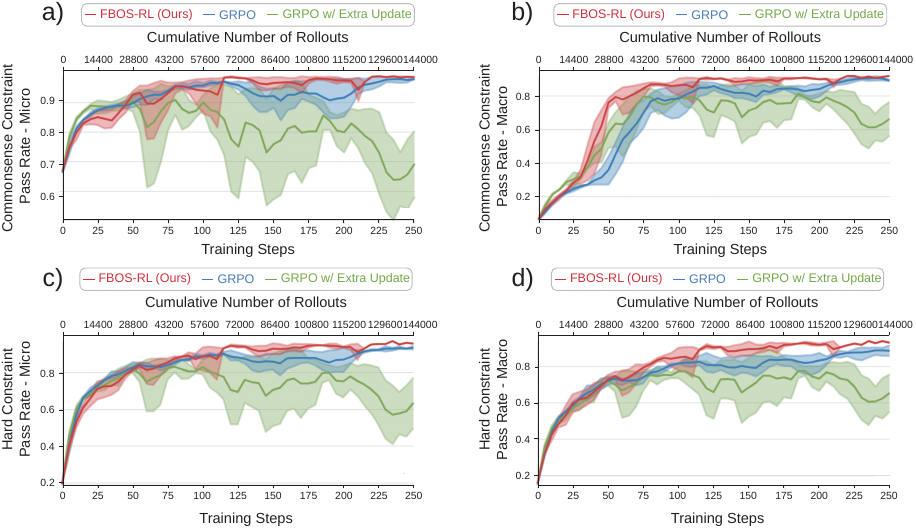}
  \vspace{-0.2in}
  \caption{
  \small
   Commonsense and hard constraint pass rates (micro and macro) of the Qwen3-14B model on the TravelPlanner validation set: our method vs.\ the GRPO w/ Extra Update baseline.
  }
  \label{fig:extra_update:constraint_pass_rate}
\end{figure}

Figure~\ref{fig:extra_update:final_pass_rate} reports the final pass rate of the Qwen3-14B model on the TravelPlanner validation set under our method and the GRPO w/ Extra Update baseline. Figure~\ref{fig:extra_update:final_pass_rate_by_difficulty} reports the final pass rate of the Qwen3-14B model across different difficulty levels. Figure~\ref{fig:extra_update:constraint_pass_rate} reports the four constraint pass rate metrics of the Qwen3-14B model on the validation set.

Figures~\ref{fig:extra_update:final_pass_rate}, \ref{fig:extra_update:final_pass_rate_by_difficulty}, and~\ref{fig:extra_update:constraint_pass_rate} indicate that our method's gains do not stem from the additional parameter update. Even when the baseline is given the same number of parameter updates per training step, our method still significantly outperforms it.

\subsubsection{Llama-3.1-8B-Instruct}
\label{app:extra_update_llama}

We further repeat the same controlled experiment on the Llama-3.1-8B-Instruct model. The experimental setup, including the construction of the GRPO w/ Extra Update baseline and the matching of the number of parameter updates per training step, is identical to that in Section~\ref{app:extra_update_qwen}.

\begin{figure}[htbp]
  \centering
  \includegraphics[width=0.6\textwidth]{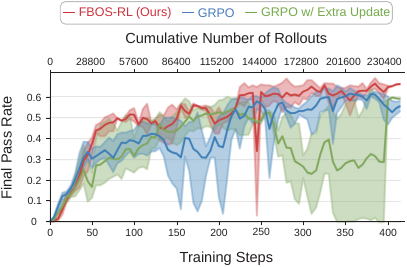}
  % \vspace{-0.2in}
  \caption{
  \small
   Final pass rate of the Llama-3.1-8B-Instruct model on the TravelPlanner validation set: our method vs.\ the GRPO w/ Extra Update baseline.
  }
  \label{fig:extra_update_llama:final_pass_rate}
\end{figure}

\begin{figure}[htbp]
  \centering
  \includegraphics[width=1\textwidth]{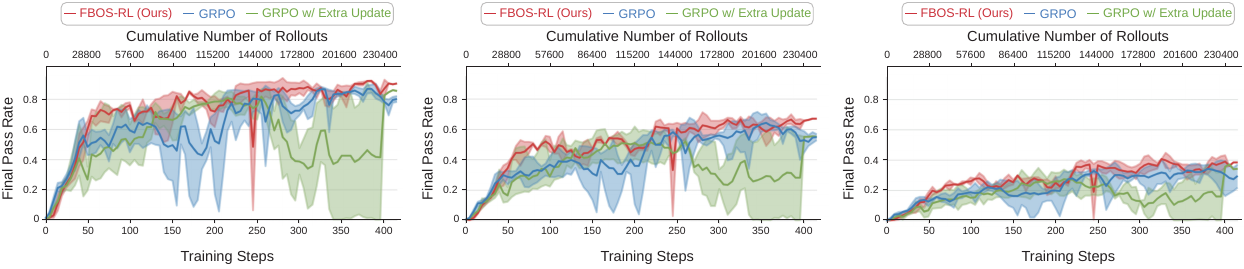}
  \vspace{-0.2in}
  \caption{
  \small
   Final pass rate of the Llama-3.1-8B-Instruct model on the TravelPlanner validation set across different difficulty levels: easy (left), medium (middle), and hard (right). Our method vs.\ the GRPO w/ Extra Update baseline.
  }
  \label{fig:extra_update_llama:final_pass_rate_by_difficulty}
\end{figure}

\begin{figure}[htbp]
  \centering
  \includegraphics[width=1\textwidth]{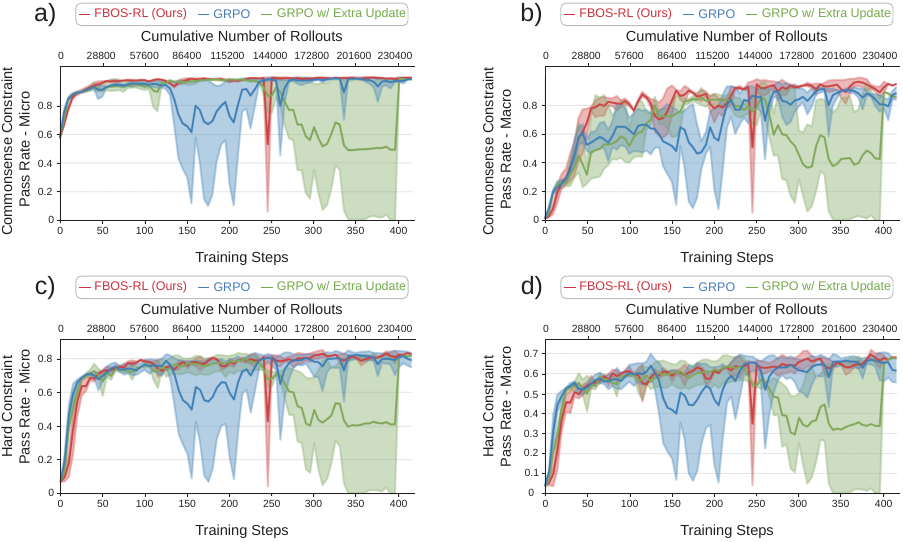}
  \vspace{-0.2in}
  \caption{
  \small
   Commonsense and hard constraint pass rates (micro and macro) of the Llama-3.1-8B-Instruct model on the TravelPlanner validation set: our method vs.\ the GRPO w/ Extra Update baseline.
  }
  \label{fig:extra_update_llama:constraint_pass_rate}
\end{figure}

Figure~\ref{fig:extra_update_llama:final_pass_rate} reports the final pass rate of the Llama-3.1-8B-Instruct model on the validation set. Figure~\ref{fig:extra_update_llama:final_pass_rate_by_difficulty} reports the final pass rate of the Llama-3.1-8B-Instruct model across different difficulty levels. Figure~\ref{fig:extra_update_llama:constraint_pass_rate} reports the four constraint pass rate metrics of the Llama-3.1-8B-Instruct model on the validation set.

Consistent with the observations on the Qwen3-14B model, Figures~\ref{fig:extra_update_llama:final_pass_rate}, \ref{fig:extra_update_llama:final_pass_rate_by_difficulty}, and~\ref{fig:extra_update_llama:constraint_pass_rate} show that our method also significantly outperforms the GRPO w/ Extra Update baseline across all metrics and difficulty levels on the Llama-3.1-8B-Instruct model, further confirming that the gains of FBOS-RL do not stem from the additional parameter update introduced by ECC.
\FloatBarrier
% \section{Conclusion}
% We present SPOGW, a score-based preference optimization method for automated agentic workflow generation that overcomes the limits of discrete optimization and pairwise comparisons via group-wise optimization in continuous space. SPOGW introduces three innovations: 1) variance-based group-wise data construction, 2) an iterative offline GRPO framework decoupling data collection from policy updates for stability, and 3) an advantage-masked KL restriction guiding policy divergence toward high-quality behaviors. Experiments on reasoning, coding, and QA benchmarks show SPOGW surpasses state-of-the-art methods, while ablations confirm each component’s contribution and highlight optimal hyperparameter settings. SPOGW offers a scalable, effective framework that reduces manual design while maintaining strong performance across domains.

% \vspace{-0.15in}
\section{Conclusion}
\label{sec:conclusion}
% \vspace{-0.1in}

We propose \textbf{FBOS-RL}, a \textbf{F}eedback-Driven \textbf{B}i-\textbf{O}bjective \textbf{S}ynergistic \textbf{R}einforcement \textbf{L}earning framework. FBOS-RL synergistically optimizes two mutually reinforcing training objectives, \emph{Exploitation-oriented Policy Alignment} (EPA) and \emph{Exploration-oriented Capability Cultivation} (ECC), forming a positive self-bootstrapping flywheel that improves both the training efficiency and the final performance ceiling of reinforcement learning.
Extensive experiments on TravelPlanner and MiniF2F-Lean4, across Llama-3.1-8B-Instruct, Qwen3-14B and Qwen3.5-27B, show that FBOS-RL substantially improves both training efficiency and the final performance ceiling over vanilla GRPO and strong controlled baselines, and exhibits clear OOD generalization to GPQA Diamond. Meanwhile, FBOS-RL avoids entropy collapse, maintains higher policy entropy, and exhibits a lower gradient norm, evidencing stronger exploration and better training stability.
%
% We hope FBOS-RL offers a new perspective on RL for LLMs: rather than treating feedback merely as data augmentation, ``understanding feedback and correcting errors'' can itself be elevated to a first-class learnable objective that co-evolves with policy alignment. Promising directions for future work include extending FBOS-RL to richer feedback sources beyond rule-based verifiers (e.g., learned critics, tool-grounded signals, or human preferences), scaling it to larger models and broader reasoning domains, and theoretically characterizing the convergence of the bi-objective flywheel.

% \vspace{-0.1in}
\paragraph{Limitations.}
FBOS-RL relies on feedback to enhance the sampling stage, and therefore its effectiveness is limited on tasks for which reliable feedback is hard to obtain.

% \section*{References}

\medskip

% \today
% ----------- 注释掉 \bibliography 命令，避免与下方粘贴的 .bbl 内容重复生成参考文献 -------------
% {\small
%  \bibliographystyle{ieee}
%  \bibliography{main}
% }

% ----------- zxk 0518 to arxiv -------------

% ----------- zxk 0518 to arxiv -------------

\appendix
\clearpage

\section{Detailed Descriptions of Datasets}
\label{app:dataset_details}

In this section, we provide the detailed descriptions of the datasets used in our experiments.

\paragraph{TravelPlanner.}
The TravelPlanner dataset is a benchmark designed to evaluate the planning capabilities of language agents in realistic travel-planning scenarios. Each query specifies a travel request (e.g., origin, destination(s), duration, budget, and personalized requirements), and the agent is required to produce a detailed multi-day itinerary.

\paragraph{MiniF2F-Lean4.}
The MiniF2F-Lean4 dataset is a benchmark for formal mathematical reasoning, consisting of Olympiad-level and high-school competition problems (drawn from sources such as AMC, AIME, and IMO) formalized as theorem-proving problems in the Lean4 proof assistant.

\section{Detailed Definitions of Evaluation Metrics}
\label{app:eval_metrics}

In this section, we provide the detailed definitions of the evaluation metrics used in the main paper, following~\cite{xie2024travelplanner}.

\begin{itemize}[leftmargin=*, parsep=0pt, itemsep=1pt, topsep=2pt]
    \item{\textit{Commonsense Constraint Pass Rate}:}
    This metric covers eight commonsense dimensions: whether the cities visited in the itinerary are reasonable, whether restaurant choices are non-repetitive, whether attractions are non-repetitive, whether the accommodations are reasonable, whether the transportation modes are reasonable, whether all daily activities (meals, attractions, accommodations) take place in the city the traveler is in on that day, whether all referenced information (e.g., restaurant names, flight numbers) exists in the closed sandbox database, and whether the plan's information is complete. It evaluates whether the model can incorporate commonsense knowledge into the generated plan without being explicitly instructed.
    \item{\textit{Hard Constraint Pass Rate}:}
    This metric measures whether the generated travel plan satisfies the hard constraints explicitly given in the query (e.g., dietary, accommodation, transportation, and budget constraints), aiming to test the model's ability to adjust its plan according to diverse user requirements.
    \item{\textit{Final Pass Rate}:}
    This metric denotes the proportion of feasible plans (i.e., those satisfying all of the aforementioned constraints, including every commonsense constraint and every hard constraint) among all evaluated plans, and measures the model's ability to generate plans that meet practical standards.
        \begin{equation}
            \text{Final Pass Rate} = \frac{\sum_{p \in P} \mathbb{I}_{\text{ IsSatisfied}(C_p^{\text{all}}, p)}}{|P|},
        \end{equation}
    where $P$ denotes the set of all plans being evaluated, and $C_p^{\text{all}}$ denotes the set of all constraints applicable to a specific plan $p$ (including all commonsense constraints and all hard constraints). $\mathbb{I}_{\text{IsSatisfied}(X, Y)}$ is an indicator function that returns 1 if $Y$ satisfies the constraint $X$, and 0 otherwise.
    For both the Commonsense Constraint Pass Rate and the Hard Constraint Pass Rate, we adopt two evaluation strategies: micro and macro.
    The micro strategy computes the proportion of satisfied constraints over the total number of constraints, and
    The macro strategy computes the proportion of plans that satisfy all commonsense constraints (or all hard constraints) among the evaluated plans,
$
    \text{Micro Pass Rate} = \frac{\sum_{p \in P} \sum_{c \in C_p} \mathbb{I}_{\text{ IsSatisfied}(c,p)}}{\sum_{p \in P} |C_p|}, \text{Macro Pass Rate} = \frac{\sum_{p \in P} \mathbb{I}_{\text{ IsSatisfied}(C_p, p)}}{|P|},
$
where $C_p$ denotes the set of constraints applicable to a specific plan $p$, and $|C_p|$ denotes the number of constraints in this set.
    These two strategies evaluate the model's ability to follow individual constraints and the full set of constraints, respectively.
    \end{itemize}

\clearpage
\section{Training Dynamics of the EPA Objective}
\label{app:epa_training_curves}

\begin{figure}[h]
    \centering
    \includegraphics[width=\linewidth]{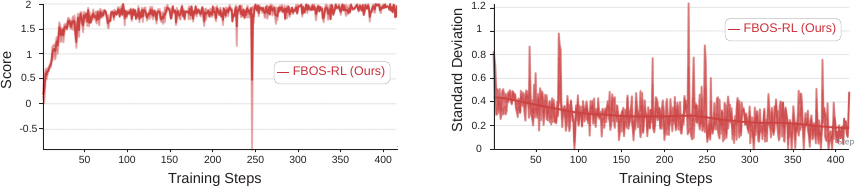}
    \caption{\small Training dynamics of the EPA objective on Llama-3.1-8B-Instruct. (a) Training-set score of EPA steadily increases along training steps. (b) The corresponding std steadily decreases, indicating that EPA can be optimized in a stable manner.}
    \label{fig:epa_llama_overall}
\end{figure}

\begin{figure}[h]
    \centering
    \includegraphics[width=\linewidth]{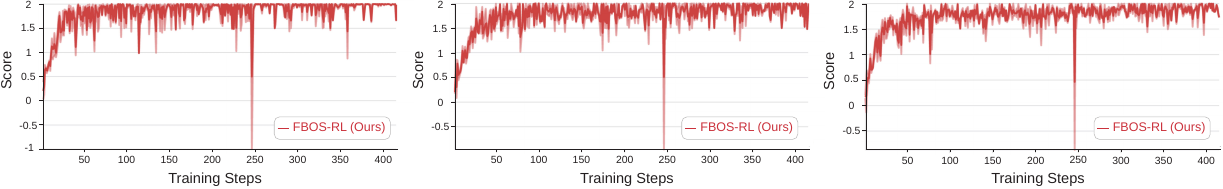}
    \caption{\small Training-set score of the EPA objective on Llama-3.1-8B-Instruct, broken down by training-sample difficulty (\textit{easy}, \textit{medium}, \textit{hard}, from left to right). The score consistently rises along training steps across all three difficulty levels.}
    \label{fig:epa_llama_difficulty}
\end{figure}

\begin{figure}[h]
    \centering
    \includegraphics[width=\linewidth]{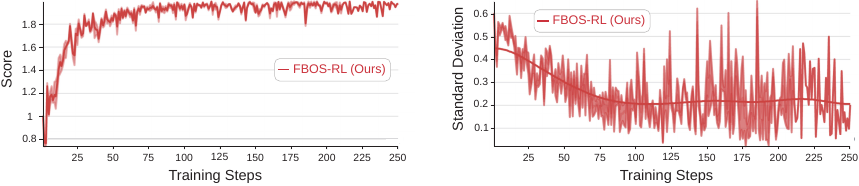}
    \caption{\small Training dynamics of the EPA objective on Qwen3-14B. (a) Training-set score of EPA steadily increases along training steps. (b) The corresponding std steadily decreases, indicating that EPA can be optimized in a stable manner.}
    \label{fig:epa_qwen_overall}
\end{figure}

\begin{figure}[h]
    \centering
    \includegraphics[width=\linewidth]{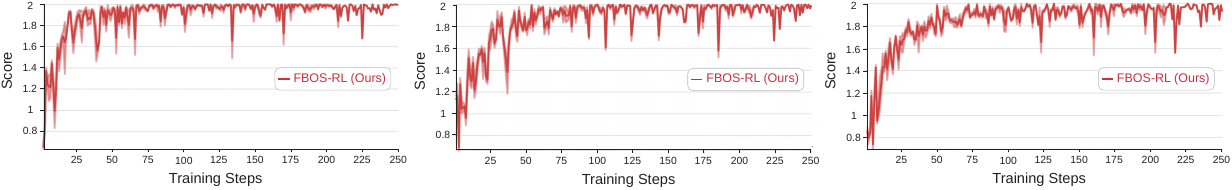}
    \caption{\small Training-set score of the EPA objective on Qwen3-14B, broken down by training-sample difficulty (\textit{easy}, \textit{medium}, \textit{hard}, from left to right). The score consistently rises along training steps across all three difficulty levels.}
    \label{fig:epa_qwen_difficulty}
\end{figure}

To empirically demonstrate that the \emph{Exploitation-oriented Policy Alignment} (EPA) objective introduced in Section~\ref{subsec:dual_objective} can be trained in a stable and effective manner, we report its training-set score and the corresponding standard deviation (std) as a function of training steps on two policy models, Llama-3.1-8B-Instruct (Figures~\ref{fig:epa_llama_overall} and~\ref{fig:epa_llama_difficulty}) and Qwen3-14B (Figures~\ref{fig:epa_qwen_overall} and~\ref{fig:epa_qwen_difficulty}). Across both models, the training-set score of EPA exhibits a clear and stable upward trend while its std steadily decreases, which together confirm that EPA can be properly optimized. Moreover, this positive trend consistently holds on training subsets of different difficulty levels (\textit{easy}, \textit{medium}, \textit{hard}), indicating that EPA is able to make robust progress regardless of sample difficulty.

\clearpage
\section{Training Dynamics of the ECC Objective}
\label{app:ecc_training_curves}

\begin{figure}[h]
    \centering
    \includegraphics[width=\linewidth]{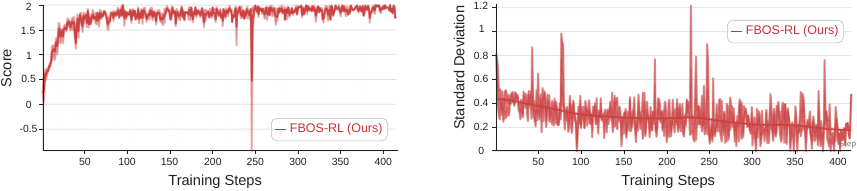}
    \caption{\small Training dynamics of the ECC objective on Llama-3.1-8B-Instruct. (a) Training-set score of ECC steadily increases along training steps. (b) The corresponding std steadily decreases, indicating that ECC can be optimized in a stable manner.}
    \label{fig:ecc_llama_overall}
\end{figure}

\begin{figure}[h]
    \centering
    \includegraphics[width=\linewidth]{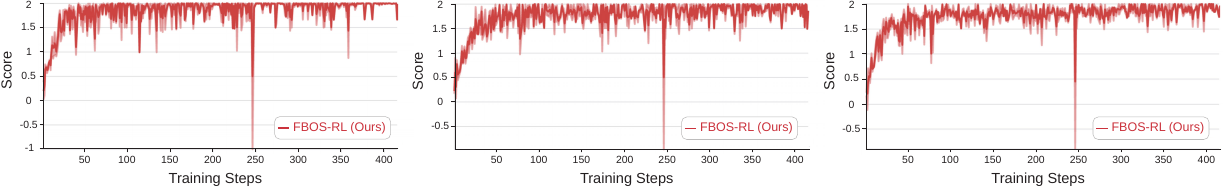}
    \caption{\small Training-set score of the ECC objective on Llama-3.1-8B-Instruct, broken down by training-sample difficulty (\textit{easy}, \textit{medium}, \textit{hard}, from left to right). The score consistently rises along training steps across all three difficulty levels.}
    \label{fig:ecc_llama_difficulty}
\end{figure}

\begin{figure}[h]
    \centering
    \includegraphics[width=\linewidth]{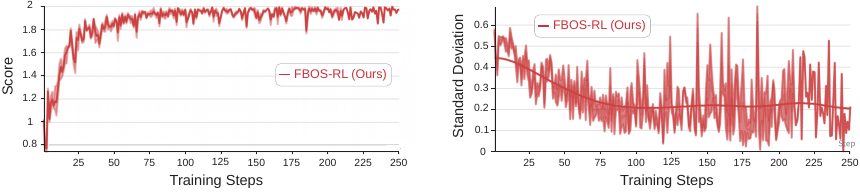}
    \caption{\small Training dynamics of the ECC objective on Qwen3-14B. (a) Training-set score of ECC steadily increases along training steps. (b) The corresponding std steadily decreases, indicating that ECC can be optimized in a stable manner.}
    \label{fig:ecc_qwen_overall}
\end{figure}

\begin{figure}[h]
    \centering
    \includegraphics[width=\linewidth]{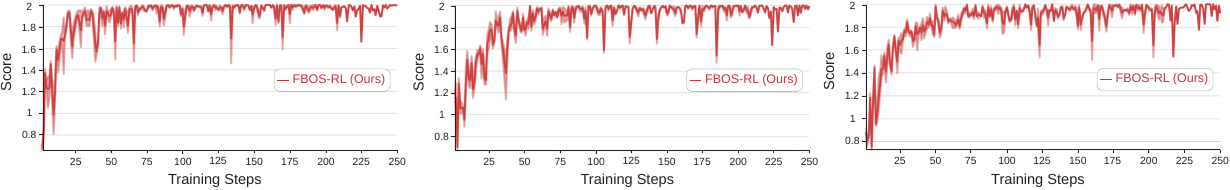}
    \caption{\small Training-set score of the ECC objective on Qwen3-14B, broken down by training-sample difficulty (\textit{easy}, \textit{medium}, \textit{hard}, from left to right). The score consistently rises along training steps across all three difficulty levels.}
    \label{fig:ecc_qwen_difficulty}
\end{figure}

We similarly validate the trainability of the \emph{Exploration-oriented Capability Cultivation} (ECC) objective introduced in Section~\ref{subsec:dual_objective}. We report the training-set score and the corresponding standard deviation (std) of ECC along training steps on Llama-3.1-8B-Instruct (Figures~\ref{fig:ecc_llama_overall} and~\ref{fig:ecc_llama_difficulty}) and Qwen3-14B (Figures~\ref{fig:ecc_qwen_overall} and~\ref{fig:ecc_qwen_difficulty}). As with EPA, we observe that the ECC training-set score steadily rises and its std steadily decreases on both models, and the same trend holds across all three difficulty levels (\textit{easy}, \textit{medium}, \textit{hard}). These results demonstrate that ECC can also be properly trained, confirming that the model indeed acquires an improved ability to discover higher-quality rollouts when guided by Feedback-Augmented Prompts (FAPs).

%%%%%%%%%%%%%%%%%%%%%%%%%%%%%%%%%%%%%%%%%%%%%%%%%%%%%%%%%%%%
% \input{NList}
\end{document}